\documentclass[10pt,twocolumn,letterpaper]{article}
\usepackage[final]{pdfpages}

\usepackage{wacv}
\usepackage{times}
\usepackage{epsfig}
\usepackage{graphicx}
\usepackage{amsmath}
\usepackage{amssymb}
\usepackage{booktabs}

\usepackage[accsupp]{axessibility}  
\usepackage{enumitem}

\usepackage{soul, color}

\usepackage{multirow}
\usepackage{tabularx}
\usepackage{makecell}
\usepackage{wrapfig}
\usepackage{nicefrac}
\usepackage[export]{adjustbox}

\usepackage{amssymb} 
\usepackage{booktabs}
\usepackage{multirow}
\usepackage{bbm}
\usepackage{bm}
\usepackage{mathtools}
\usepackage{xfrac}
\usepackage{lipsum}  
\usepackage[sectionbib,numbers,sort,compress]{natbib}
\usepackage{chapterbib}

\newcommand{\nsparagraph}[1]{\noindent \textbf{#1}}
\newcommand{\lsparagraph}[1]{\vspace{0.1cm} \noindent \textbf{#1}}
\newcommand{\msparagraph}[1]{\vspace{0.3cm} \noindent \textbf{#1}}

\DeclareMathOperator*{\argmin}{arg\,min} 

\usepackage{colortbl}
\usepackage{pifont}

\newcommand{\xmark}{{\ding{55}}}

\usepackage{xcolor}
\definecolor{cb-black}      {RGB}{  0,   0,   0}
\definecolor{cb-blue-green} {RGB}{  0,  073,  073}
\definecolor{cb-green-sea}  {RGB}{  0, 146, 146}
\definecolor{cb-rose}       {RGB}{255, 109, 182}
\definecolor{cb-salmon-pink}{RGB}{255, 182, 119}
\definecolor{cb-purple}     {RGB}{ 73,   0, 146}
\definecolor{cb-blue}       {RGB}{ 0, 109, 219}
\definecolor{cb-lilac}      {RGB}{182, 109, 255}
\definecolor{cb-blue-sky}   {RGB}{109, 182, 255}
\definecolor{cb-blue-light} {RGB}{182, 219, 255}
\definecolor{cb-burgundy}   {RGB}{146,   0,   0}
\definecolor{cb-brown}      {RGB}{146,  73,   0}
\definecolor{cb-clay}       {RGB}{219, 209,   0}
\definecolor{cb-green-lime} {RGB}{ 36, 255,  36}
\definecolor{cb-yellow}     {RGB}{255, 255, 109}

\colorlet{light_grey}{gray!15}
\sethlcolor{light_grey}

\usepackage[pagebackref,colorlinks,breaklinks=true,bookmarks=false]{hyperref}
\usepackage[capitalize]{cleveref}
\crefname{section}{Sec.}{Secs.}
\Crefname{section}{Section}{Sections}
\Crefname{table}{Table}{Tables}
\crefname{table}{Tab.}{Tabs.}

\def\se3{\mathbf{SE}(3+)}
\def\SE3{\mathbf{SE}(3)}
\def\SO3{\mathbf{SO}(3)}

\setboolean{@twoside}{false}

\def\wacvPaperID{1309} 
\wacvfinalcopy 

\begin{document}

\title{Self-supervised Correspondence Estimation via Multiview Registration}

\author{
Mohamed El Banani$^{1}$\thanks{~Work done during an internship at Meta AI.}  \\
\tt\small mbanani@umich.edu
\and
Ignacio Rocco$^{2}$ \\
\tt\small irocco@meta.com
\and
David Novotny$^{2}$ \\
\tt\small dnovotny@meta.com
\and
Andrea Vedaldi$^{2}$ \\
\tt\small vedaldi@meta.com
\and
Natalia Neverova$^{2}$ \\
\tt\small nneverova@meta.com
\and
Justin Johnson$^{1,2}$ \\
\tt\small justincj@umich.edu
\and
Ben Graham$^{2}$ \\
\tt\small benjamingraham@meta.com\and
{\large $^1$University of Michigan \qquad $^2$Meta AI}
}

\maketitle
\thispagestyle{empty}

\begin{abstract}
Video provides us with the spatio-temporal consistency needed for visual learning. 
Recent approaches have utilized this signal to learn correspondence estimation from close-by frame pairs. 
However, by only relying on close-by frame pairs, those approaches miss out on the richer long-range consistency between distant overlapping frames. 
To address this, we propose a self-supervised approach for correspondence estimation that learns from multiview consistency in short RGB-D video sequences. 
Our approach combines pairwise correspondence estimation and registration with a novel SE(3) transformation synchronization algorithm.
Our key insight is that self-supervised multiview registration allows us to obtain correspondences over longer time frames; increasing both the diversity and difficulty of sampled pairs. 
We evaluate our approach on indoor scenes for correspondence estimation and RGB-D pointcloud registration and find that we perform on-par with supervised approaches.

\end{abstract}

\begin{figure}[t]
\begin{center}
\includegraphics[width=0.95\linewidth]{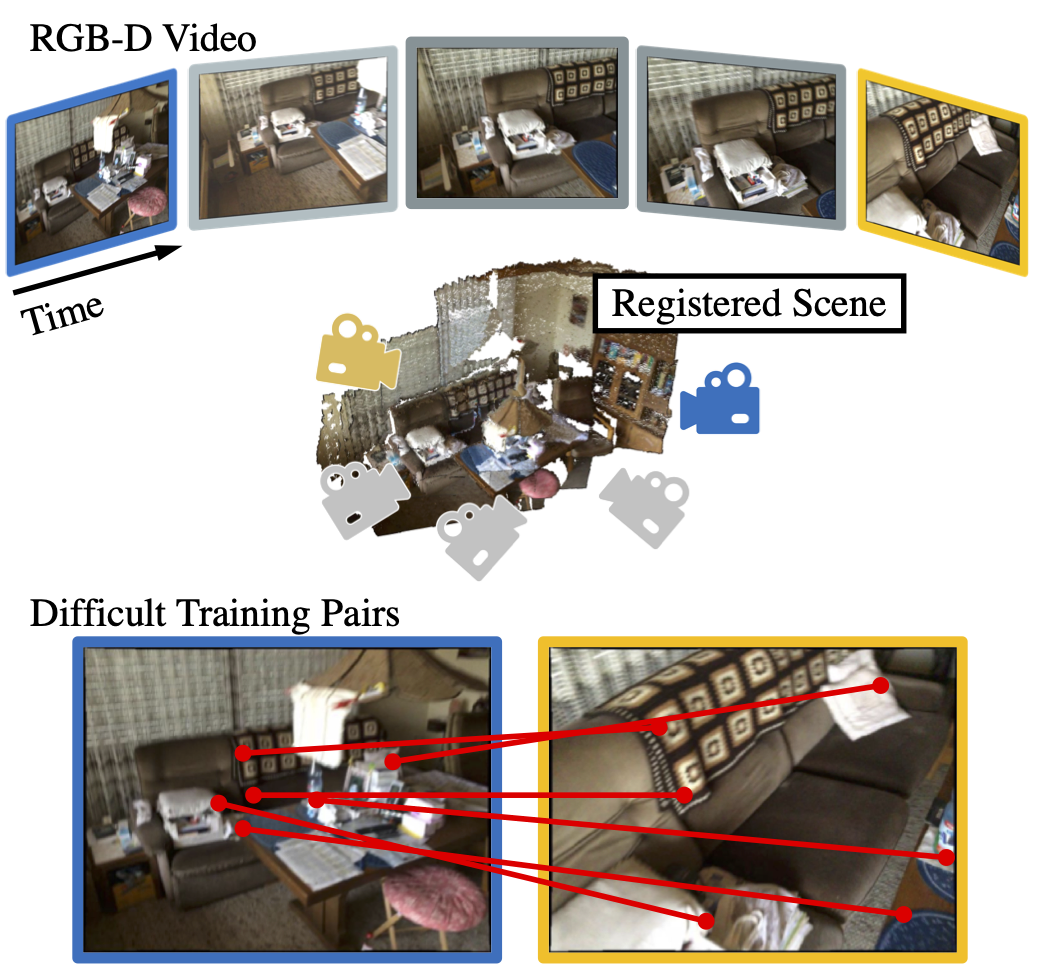}
\vspace{-2mm}
\end{center}
\caption{
Multiview RGB-D registration allows us to learn from wide-baseline view pairs which exhibit more appearance change that adjacent frames.
}
\vspace{-2mm}
\label{fig:teaser}
\end{figure}

\section{Introduction}
\label{sec:intro}
Consider the couch in \cref{fig:teaser}. 
While the start and end frames depict overlapping regions in space, 
the large viewpoint change makes them appear significantly different.
The ability to establish correspondence across views lies at the core of scene understanding and visual tasks such as SLAM and structure-from-motion. 
The common approach to learning correspondence estimation relies on correspondence supervision; \ie, telling the model which pixels belong to the same point in space. 
However, we commonly learn about the world by moving and observing how appearance changes without such explicit supervision.  
\textit{Could we learn correspondence estimation directly from video?}

Modern correspondence estimation approaches rely heavily on supervision. 
This is typically obtained by applying classical 3D reconstruction algorithms on large image collections~\cite{snavely2006photo,wilson20141dsfm,MegaDepthLi18} or carefully captured indoor scans~\cite{dai2017scannet,dai2017bundlefusion,chang2017matterport3d}, and then sampling overlapping view pairs for training. 
This has been widely successful, as it provides correspondence supervision with large viewpoint and lighting changes.
While current methods benefit from supervision, they are limited to learning from carefully captured videos that can already be constructed using standard algorithms. 
Recently, there has been a rise in self-supervised correspondence approaches that rely on close-by frames in video~\cite{elbanani2021unsupervisedrr,elbanani2021byoc,jabri2020walk,lai2019sslvidcorrflow}. 
This sampling strategy limits the appearance change, as shown in \cref{fig:teaser}, resulting in poor performance on image pairs with large viewpoint changes. 
Ideally, we would leverage the temporal consistency within the video to learn from distant overlapping frames while ignoring non-overlapping pairs. 

To this end, we propose SyncMatch: a self-supervised approach for learning correspondence estimation through synchronized multiview pointcloud registration.
Our approach bootstraps itself, generating wide-baseline view pairs through registering and synchronizing all pairwise transformations within short RGB-D video clips. 
Our core insight is that through synchronizing transformations across longer time frames, we can detect and learn from difficult pairs with large viewpoint changes. Despite only relying on geometric consistency within RGB-D videos, we achieve comparable performance to fully-supervised approaches. 

Our approach is inspired by self-supervised pointcloud registration~\cite{elbanani2021unsupervisedrr,elbanani2021byoc} and transformation  synchronization~\cite{arrigoni2016spectral,gojcic2020learning}. 
The core insight in self-supervised pointcloud registration is that randomly initialized networks provide sufficiently good features for narrow-baseline pointcloud registration.
This allows them to provide good pseudolabels for self-supervised training.
Meanwhile, transformation synchronization allows us to estimate an accurate camera trajectory from a potentially noisy set of pairwise relative camera poses. 
Our approach combines both ideas for self-supervised multiview registration; allowing us to learn correspondence estimation across large viewpoint changes. 

We evaluate our approach on RGB-D indoor scene videos. 
We train our model on RGB-D videos from ScanNet and ETH-3D, and evaluate it on correspondence estimation and RGB-D pointcloud registration. 
Despite only learning from RGB-D video, our approach achieves a similar performance to supervised approaches with more sophisticated matching algorithms. Furthermore, we provide a comprehensive analysis of our approach to understand the impact of the training data and the model components. 

In summary, our contributions are as follows:
\begin{itemize}[noitemsep,nolistsep,leftmargin=*]
    \item A self-supervised correspondence estimation approach based on multiview consistency in RGB-D video.
    \item A novel SE(3) transformation synchronization algorithm that is fast, and numerically stable during training.
\end{itemize}

\section{Related Work}
\label{sec:related}
\nsparagraph{Correspondence Estimation.}
Correspondence estimation is the task of identifying points in two images that correspond to the same physical location.
The standard approach for establishing correspondence has two distinct steps: feature extraction and feature matching. 
Early work exploited hand-crafted feature detectors~\cite{lowe2004distinctive,mikolajczyk2004scale} to extract normalized image patches across repeatable image points, combined with hand-crafted descriptors based on local statistics to obtain features with some robustness to illumination changes and small translations~\cite{lowe2004distinctive,arandjelovic2012three, bay2006surf}.
These features are matched via nearest neighbor search and filtered using heuristics; \eg ratio test~\cite{lowe2004distinctive} or neighborhood consensus~\cite{schmid1997local}.
With the advent of deep learning, learnt keypoint detectors~\cite{lenc2016learning,laguna2019key}, descriptors~\cite{tian2017l2, balntas2016learning}, and correspondence estimators~\cite{yi2018learning,sarlin2020superglue,rocco2018neighbourhood} have been proposed.
These models are trained using correspondence supervision from traditional 3D reconstruction algorithms~\cite{schonberger2016structure,dai2017bundlefusion} on 
image collections of tourist landmarks~\cite{snavely2006photo,wilson20141dsfm,MegaDepthLi18} or indoor scene video scans~\cite{chang2017matterport3d,dai2017scannet,zeng20163dmatch}. 
Other approaches have explored self-supervision using synthetic data~\cite{zhou2017unsupervised, rocco2017convolutional, detone2018superpoint,melekhov2021digging}, traditional descriptors~\cite{yang2021self}, or RGB-D pairs from video~\cite{elbanani2021byoc,elbanani2021unsupervisedrr}. 
Our work shares the motivation of self-supervised approaches and extends it to learning from multiview consistency to better exploit the rich signal in video. 

\nsparagraph{Pointcloud Registration. }
Pointcloud registration is the task of finding a transformation that aligns two sets of 3D points. 
Early work assumes access to perfect correspondence and devised algorithms to estimate the rigid body transformation that would best align them~\cite{kabsch1976solution,umeyama1991least,arun1987least}. Later work proposed robust estimators that can handle correspondence errors and outliers that arise from feature matching~\cite{zhang1995robust,fischler1981ransac}.
More recently, learned counterparts have been proposed for 3D keypoint descriptors~\cite{choy2019fully,deng2019directreg,graham2015sparse3dcnn}, 
correspondence estimation~\cite{brachmann2017dsac, brachmann2019neural,choy2020deep,huang2020featureregistration,gojcic2020learning,ranftl2018deepfundamental,yew2020RPMNet}, as well as models for directly registering a pair of pointclouds~\cite{deng2019directreg,le2019sdrsac,lu2019deepvcp,hertz2020pointgmm}.
Closest to our work are approaches that use RGB-D video to learn correspondence-based pointcloud registration~\cite{elbanani2021unsupervisedrr,elbanani2021byoc,wang2022improving}.
Similar to our approach, they learn from the geometric consistency between RGB-D frames in a video. However, unlike those approaches, we train on short sequences of videos instead of frame pairs, allowing us to train on view pairs with larger camera changes.

\nsparagraph{SE(3) Transformation Synchronization} 
Given a set of pairwise estimates, synchronization estimates that set of latent values that explains them. 
Transformation synchronization refers to this problem applied to SO(3) and SE(3) transformations, as it commonly arises in SLAM settings~\cite{kummerle2011g,carlone2015initialization,mur2015orb,salas2013slam++}. 
For video, one could naively only consider adjacent pairs to construct a minimum spanning tree and aggregate the transformations. 
However, this only works if all pairwise estimates are accurate, since a single bad estimate can be detrimental.
More robust approaches have been proposed that can leverage the information from multiple (or all) edges in the
graph~\cite{huber2003fully,birdal2020measure,kummerle2011g,huang2019learning,huang2017translation,roessle2022end2end}.
Most relevant to our work is Arrigoni~\etal~\cite{arrigoni2014robust,arrigoni2016spectral} and Gojcic~\etal~\cite{gojcic2020learning}. 
Arrigoni~\etal~\cite{arrigoni2014robust,arrigoni2016spectral} propose a closed-form solution to SO(3) and SE(3) synchronization based on the eigendecomposition
of a pairwise transformation matrix. 
Gojcic~\etal~\cite{gojcic2020learning} builds on those ideas and integrates transformation synchronization with a supervised end-to-end pipeline for multiview registration.
We are inspired by this work and propose a different approach to SE(3) synchronization based on iterative matrix multiplication, which allows for accurate
synchronization while being more numerically stable. 
Furthermore, unlike prior work~\cite{gojcic2020learning,huang2019learning,roessle2022end2end}, we use transformation synchronization for learning without supervision.

\begin{figure*}[t!]
\begin{center}
\includegraphics[width=\textwidth]{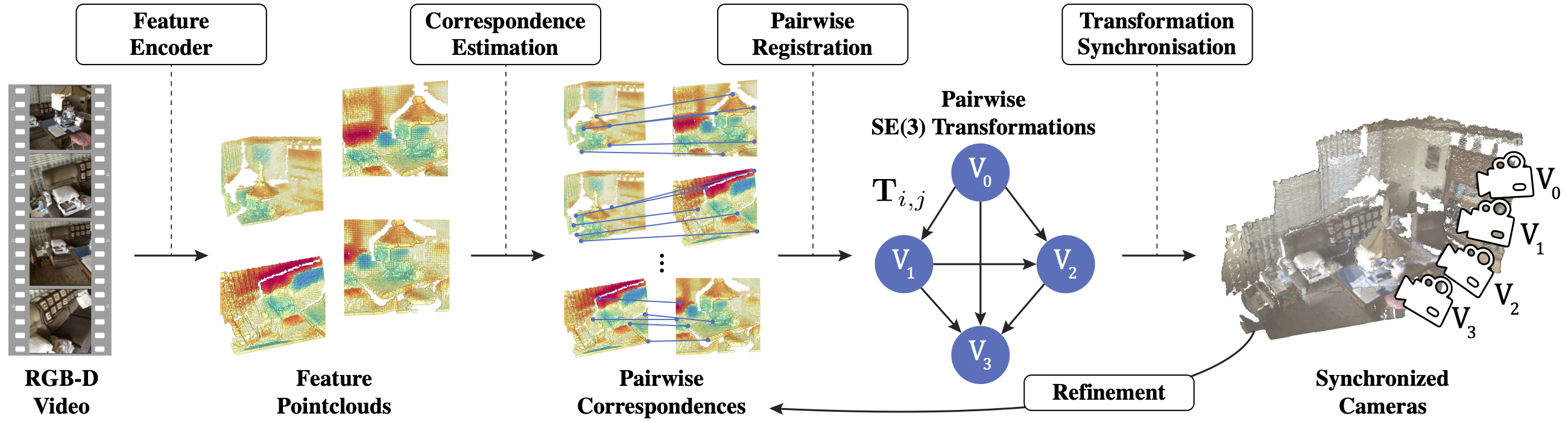}
\end{center}
\vspace{-0.2cm}
\caption{
\textbf{SyncMatch.} Given a sequence of $N$ RGB-D video frames, we extract features for each image and project them to a 3D pointcloud using input depth. 
We extract all pairwise correspondences to estimate pairwise SE(3) transformations.
We then synchronize the pairwise transformations to register the scene.
Finally, we use the estimated registration to refine our correspondence and transformation estimates.
We compute correspondences losses for both the initial and refined registrations, and backpropagate them to the feature encoder. 
}
\label{fig:model}
\vspace{-0.2cm}
\end{figure*}

\section{Approach}
\label{sec:approach}

We learn correspondence estimation from multiview registration of short RGB-D sequences without relying on pose or correspondence supervision.
We first provide a high-level sketch of our approach, shown in \cref{fig:model}, before discussing each component in detail. 

\lsparagraph{Approach Sketch.} 
Given $N$ RGB-D frames, we extract features for each RGB image and project them onto a 3D pointcloud using input depth and camera intrinsics. 
We then extract correspondences between all pointcloud pairs and estimate pairwise SE(3) transformations. 
Given $\binom{N}{2}$ pairwise transformations, we apply transformation synchronization to find the $N$ camera extrinsic parameters in a shared global frame. 
Given this coarse alignment, we resample correspondences based on both feature and spatial proximity. We repeat the registration using the updated correspondences. 
Finally, we compute the loss using the estimated correspondences and SE(3) transformations and backpropagate it to the feature encoder.

\subsection{Feature Pointcloud}
\label{sec:approach_features}

We use a randomly-initialized ResNet-18 for feature extraction.
While correspondence estimation methods often rely on keypoint detection, our approach is detector free and generates dense features uniformly across the image.
Similar to Sun~\etal~\cite{sun2021loftr}, we generate the feature grid at a lower resolution ($\sfrac{1}{4}$) than the input image. 
For each frame $i$, we use the input depth map and camera intrinsics to project the features into a pointcloud  $\mathcal{P}_i$ where each point $p \in \mathcal{P}_i$ has a feature $\mathbf{f}_p$ and a 3D coordinate $\mathbf{x}_p$.

\subsection{Correspondence Estimation}
\label{sec:approach_correspondence}

We estimate feature correspondences for all view pairs $(i, j)$. 
We first generate an initial set of correspondences by finding for each point in image $i$ the point in image $j$ with the closest matching feature vector.
The initial correspondence set will include mismatches due to poor matching, repetitive textures, or occlusion.

Correspondences can be filtered using measures of unique matches or geometric consistency. 
Heuristics such as Lowe's ratio test~\cite{lowe2004distinctive} prefer unique matches and have been extended to self-supervised pointcloud registration~\cite{elbanani2021byoc,elbanani2021unsupervisedrr} and attention-based matching~\cite{huang2021predator,sarlin2020superglue,sun2021loftr}. 
Geometric consistency relies on the idea a geometrically consistent set of correspondences is likely correct. 
This can be done by estimating the transformation similar to RANSAC~\cite{fischler1981ransac} or directly estimating the inlier scores~\cite{choy2020deep,yi2018learning,ranftl2018deepfundamental}.
We use the ratio test for initial alignment and leverage geometric consistency for refinement (\cref{sec:approach_refinement}). 
Specifically, we compute a ratio between the cosine distances in feature space as follows:
\begin{equation}
w_{p, q} =  1 - \frac{D(p, q)}{D(p, q')},
\end{equation}
where $D(p, q)$ is the cosine distance between the features, and $q$ and $q'$ are the first and second nearest neighbors to point $p$ in feature space. 
We use the weights to rank the correspondences and only keep the top $k$ correspondences. This results in a correspondence set $\mathcal{C}_{i,j}$ for each pair of frames. The correspondences $(p, q, w_{p, q})\in \mathcal{C}_{i, j}$ consists of the two matched points and the match weight.

\subsection{Pairwise Alignment}
\label{sec:approach_alignment}
For each pair of frames, we can identify a transformation $\mathbf{T}_{i,j} \in \text{SE(3)}$ that minimizes the weighted mean-squared error between the aligned correspondences across the images:
\begin{equation}
\mathbf{T}_{i,j} = \argmin_{\mathbf{T} \in \text{SE}(3)} \sum_{(p, q, w) \in \mathcal{C}_{i,j}} w||\mathbf{x}_{q} - \mathbf{T}(\mathbf{x}_p)||_2^2.
\end{equation}
A differentiable solution is given by the Weighted Procrustes (WP) algorithm~\cite{choy2020deep} which adapts the classic Umeyama pointcloud alignment  algorithm~\cite{kabsch1976solution,umeyama1991least}.

\lsparagraph{WP-RANSAC. }
While the WP algorithm can handle small errors, it is not robust against outliers.  
El Banani~\etal~\cite{elbanani2021unsupervisedrr} propose combining the alignment algorithm with random sampling to increase robustness. 
However, a single large outliers can still perturb the solution since solutions are still ranked according to the average residual error on all matches. 
We modify the WP algorithm to more closely resemble classic RANSAC~\cite{fischler1981ransac}.
We randomly sample $k$ correspondence subsets, estimate $k$ transformations, and compute an inlier score based on each transformation. 
We choose the transformation that maximizes the inlier score instead of minimizing the weighted residual error~\cite{elbanani2021unsupervisedrr}; making us more robust to large outliers. 
Finally, we update the correspondence weights with the inlier scores which can zero out large outliers. 
We compute the final registration using the WP algorithms with the updated weights; maintaining differentiability with respect to the correspondence weights.

\subsection{SE(3) Transformation Synchronization}
\label{sec:approach_synchronization}

Given estimates for the $\binom{N}{2}$ camera-to-camera transformations, we want to find the $N$ world-to-camera transformations that best explain them.
Arrigoni~\etal~\cite{arrigoni2014robust,arrigoni2016spectral} propose a closed-form solution to SE(3) synchronization using spectral decomposition. This approach was latter extended to end-to-end learning pipelines~\cite{huang2019learning,gojcic2020learning}.
The approach operates by constructing a block matrix of pairwise transformations where block ($i, j$) corresponds to the transformation between camera $i$ to camera $j$. 
The core insight in that line of work is that the absolute transformations constitute the basis of the pairwise transformations matrix, and hence, can be recovered using eigendecomposition. 

While those approaches are successful for inference, they suffer from numerical instabilities during training.
This is caused by the backward gradient of the eigendecomposition scaling 
with $\frac{1}{\text{min}_{i\neq j} \lambda_i - \lambda_j}$ where $\lambda$ are the eigenvalues.
Given that the rank of a perfect SE(3) pairwise matrix is 4, $\text{min}_{i\neq j} \lambda_i - \lambda_j$ approaches 0 for an accurate pairwise matrix which results in exploding gradients.
We observed this instability during training. 
To avoid this instability, we compute the relevant part of the eigendecomposition by power iteration, similar to PageRank~\cite{page1999pagerank}.
We observe that this converges quickly while being stable during training. 
We refer the reader to the appendix for more details.

\begin{figure}[t!]
\centering
\includegraphics[width=0.9\linewidth]{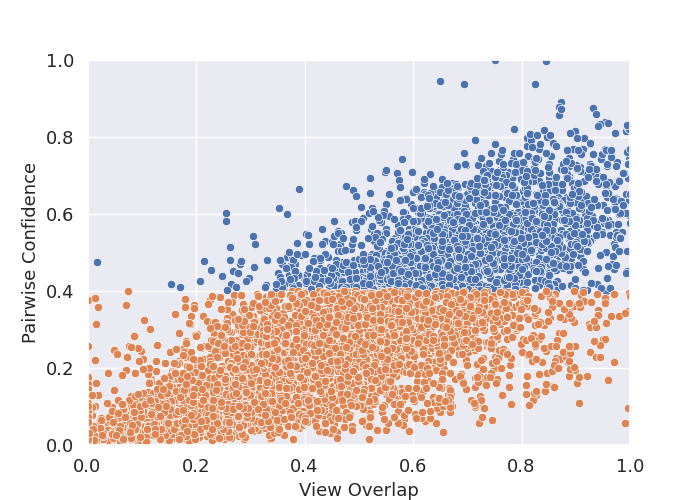}
\caption{\textbf{Mean correspondence confidence is an effective overlap filter.} 
While confidence is not perfectly correlated with view overlap, simple thresholding can accurately filter out low- and no-overlap view pairs.}
\label{fig:overlap_confidence}
\end{figure}

\lsparagraph{Pairwise confidence.}
Synchronization is more effective when provided with confidence weights for each of the pairwise estimates. 
While prior approaches train separate networks to estimate pairwise confidence~\cite{huang2019learning,gojcic2019perfect}, we instead opted for a simpler approach.
We observe that the mean confidence weight is well correlated with view overlap as shown in~\cref{fig:overlap_confidence}, where pairwise confidence is computed as $\hat{c}_{i, j}  = \frac{1}{|\mathcal{C}_{i, j}|}\sum\limits_{w \in \mathcal{C}_{i, j}} w.$

While confidence is correlated with view overlap, non-overlapping frames still receive a non-zero confidence. 
Incorrect transformations from non-overlapping pairs can negatively affect both synchronization and learning.
We address this by rescaling the confidence values based on a threshold to ignore any non-overlapping pairs. 
While this criteria is simple, it has many false negatives as shown in~\cref{fig:overlap_confidence}. 
To ensure that synchronization is always possible, we exclude adjacent pairs from rescaling since we know they most likely overlap. 
The pairwise confidence terms are adjusted as follows:
\begin{equation}
c_{i, j} = 
\begin{cases} 
      \text{max}(0,~\hat{c}_{i, j} - \gamma)\big/(1 - \gamma) & \text{if $|i - j| > 1$},\\[2pt]
      \hat{c}_{i, j} & \text{otherwise}. 
\end{cases}
\end{equation}
where $\gamma$ is the confidence threshold. 
We only estimate pairwise correspondences for pairs $(i, j)$ where $i < j$ to avoid repeated calculations. 
For pairs where $j > i$, we set $c_{j, i} = c_{i, j}$ and $T_{j, i} = T_{i, j}^{-1}$. 

\lsparagraph{Pairwise Transformation Matrix. }
We form a block matrix $\mathbf{A}$ using the weighted transformations as follows:
\begin{equation}
    \mathbf{A}  = \begin{bmatrix*}[c] 
    c_{1} \mathbf{I}_4 & c_{1, 2}\mathbf{T}_{1, 2} & \cdots & c_{1, N}\mathbf{T}_{1, N}\\
    c_{2, 1}\mathbf{T}_{2, 1} & c_{2} \mathbf{I}_4 & \cdots & c_{2, N}\mathbf{T}_{2, N}\\
    \vdots & \ddots &  & \vdots\\
    c_{N, 1}\mathbf{T}_{N, 1} &  c_{N, 2}\mathbf{T}_{N, 2} & \cdots & c_{N} \mathbf{I}_4 \\
\end{bmatrix*},
\end{equation}
where $c_{i, j}$ is the pairwise confidence, $\mathbf{T}_{i, j}$ is the estimated pairwise transformation, and $c_{i} = \sum_{k \in N} c_{i, k}$. We perform $t$ matrix multiplications to calculate $\mathbf{A}^{2^t}$ and extract the synchronized transformations by taking the first block column and normalizing each transformation by its confidence (bottom right element). 
This results in $N$ SE(3) transformations in the first view's frame of reference. 

\begin{figure}[t!]
\begin{center}
\includegraphics[width=0.9\linewidth]{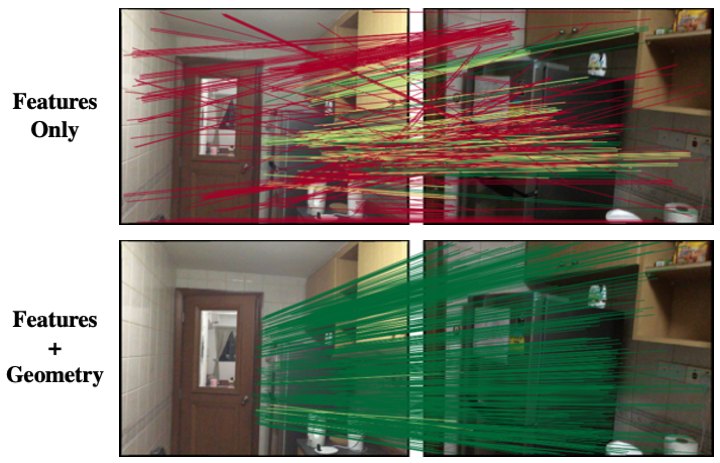}
\end{center}
\caption{\textbf{Geometry-aware sampling greatly improves our correspondence set.} 
GART allows us to extract accurate correspondence even if the initial feature matches are very noisy. }
\label{fig:refinement}
\end{figure}

\subsection{Refinement}
\label{sec:approach_refinement}

While feature-based matching is powerful, it can produce false positive feature matches that could be easily filtered out through geometric consistency.
To this end, we use the predicted scene alignment to refine our correspondences by filtering matches that are not geometrically consistent with the estimated transformation. 
We resample the correspondences but compute the ratio test based on both feature similarity and spatial proximity. We update our correspondence filtering criteria by changing the distance function in the ratio test to:
\begin{equation}
    D_{refine}(p, q) = D_{C}(\mathbf{f}_p, \mathbf{f}_{q}) + \lambda \|\mathbf{x}_p - \mathbf{x}_q \|_2,
\end{equation}
where $D_{C}(x, y)$ is cosine distance, $\mathbf{f}_p$ is the feature vector belonging to point $p$, $\lambda$ is a weighing constant, and $\mathbf{x}_p$ is the aligned 3D coordinate of point $p$ in a common global frame. We refer to this updated ratio test as a Geometry-Aware Ratio Test (GART).

\subsection{Losses}
\label{sec:approach_losses}

We emphasize that our approach is self-supervised. Hence, we only rely on the consistency within the video for training.
We use the registration loss proposed by El Banani and Johnson~\cite{elbanani2021byoc} which minimizes the weighted residual error of the estimated correspondences using the estimated alignment. For a given pair $(i, j)$, we compute a registration loss as follows:  
\begin{equation}
\mathcal{L}_{reg}(i, j) = \sum_{(p, q, w) \in \mathcal{C}_{i, j}} w\|\mathbf{T}^{-1}_{j}\mathbf{x}_{q} - \mathbf{T}^{-1}_{i}\mathbf{x}_p\|_2,
\end{equation}
where $p$ and $q$ are the corresponding points, $w$ is their weight, and $\mathbf{T}_{i}$ is the synchronized transformation (\cref{sec:approach_synchronization}).
We compute this loss for both the initial and refined correspondence sets and predicted transformations for all view pairs.

\section{Experiments}
\label{sec:experiments}
We evaluate our approach on two indoor scene datasets: ScanNet~\cite{dai2017scannet} and ETH3D~\cite{schoeps2017cvpr}. 
Our experiments address the following questions: 
(1) does multiview training improve over pairwise training?;
(2) can multiview self-supervision replace full-supervision?; 
(3) can we reconstruct scenes from RGB-D sequences?;
(4) can we learn from videos that cannot be reconstructed using standard approaches?

\nsparagraph{Training Dataset.} 
ScanNet provides RGB-D videos of 1513 scenes and camera poses computed using BundleFusion~\cite{dai2017bundlefusion}.
We use the train/valid/test scene split from ScanNet v2.
While automated reconstruction models like BundleFusion are able to reconstruct ScanNet scenes, ETH3D~\cite{schoeps2017cvpr} videos are more challenging to such systems with BundleFusion failing on most of those sequences.
As a result, ETH3D offers us with an interesting set of videos which cannot be currently used by supervised training. 

We emphasize that we only use RGB-D video and camera intrinsics for training and any provided camera poses are only used for evaluation.
We generate view pairs by sampling views that are 20 frames apart. For longer sequences, we combine adjacent pairs to get N-tuples.  

\nsparagraph{Training Details. }
We train our model with the AdamW~\cite{kingma2015adam,loshchilov2018decoupled} optimizer using a learning rate of $10^{-3}$ and a weight decay of $10^{-3}$.
We train for 100K iterations with a batch size of 16. 
Unless otherwise stated, we use 6 views for training. 
Our implementation is in PyTorch~\cite{paszkepytorch}, with extensive use of PyTorch3D~\cite{pytorch3d}, FAISS~\cite{FAISS}, PyKeOps~\cite{PyKeOps}, and Open3D~\cite{open3d}. We make our code publicly available.\footnote{\url{https://github.com/facebookresearch/SyncMatch}}

\begin{figure*}[t!]
\begin{center}
\includegraphics[width=0.98\textwidth]{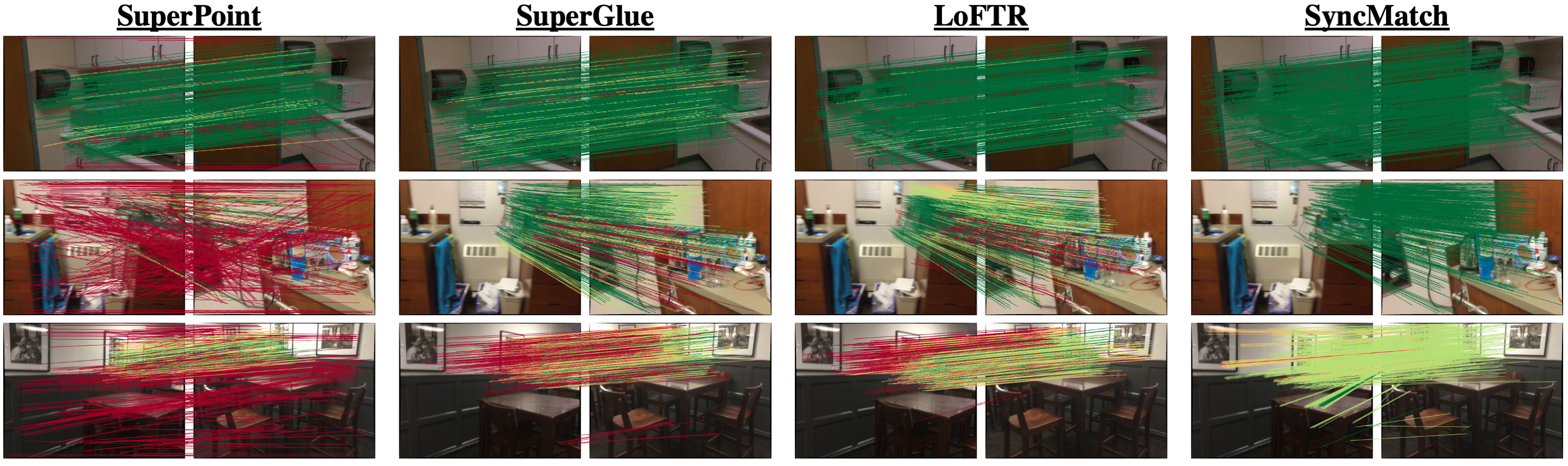}
\vspace{-10pt}
\end{center}
\label{fig:qualitiative_corr}
\caption{\textbf{Correspondence Estimation on ScanNet.}
Our model extracts accurate correspondences for large viewpoint change.
Through combining both strong feature descriptors and geometric refinement, we can successfully handle cases where prior approaches fail.
Correspondences are color-coded by 3D error. 
}
\vspace{-8pt}
\end{figure*}

\subsection{Correspondence Estimation}

We evaluate our model on correspondence estimation. 
While our model is trained on adjacent pair sequences, the primary challenge is how it performs for wide-baseline correspondence estimation. 
We evaluate this on the test set proposed by SuperGlue~\cite{sarlin2020superglue} of 1500 view pairs. 
This dataset includes difficult pairs with large camera motions; \eg, images from opposite sides of a room. 

\lsparagraph{Evaluation Metrics.}
We evaluate the estimated correspondences based on their 2D and 3D errors.
We lift the estimated correspondences into 3D using known depth and intrinsics and only consider keypoints with valid depth values. 
We use groundtruth transformations to align the keypoints and compute the 3D error and the 2D reprojection error. 
We extract 500 correspondences for all methods to allow for a meaningful comparison between precision values.\footnote{LoFTR and SuperGlue use a mutual check heuristic for matching which can produce fewer correspondences. In such cases, we use all the correspondences they produce.}

\lsparagraph{Baselines.}
We compare our approach against classic, self-supervised, and supervised correspondence estimation methods.
First, we compare against two commonly used feature descriptors: RootSIFT~\cite{arandjelovic2012three} and SuperPoint~\cite{detone2018superpoint}. 
RootSIFT is a hand-crafted feature that is still used in modern pipelines, while
SuperPoint is a self-supervised descriptor trained on synthetic data and affine transformations of web images.
We report the performance of these features for image-only matching using the ratio test~\cite{lowe2004distinctive} as well as depth-refined matching using our proposed GART. 

We consider three end-to-end approaches: SuperGlue~\cite{sarlin2020superglue}, LoFTR~\cite{sun2021loftr}, and BYOC~\cite{elbanani2021byoc}. 
SuperGlue is an attention-based matching algorithm built on top of SuperPoint. 
LoFTR and BYOC are both detector-free approaches which train their own features from scratch.
SuperGlue and LoFTR both use transformers for matching that are trained with correspondence supervision. 
BYOC is self-supervised and uses a variant of the ratio test for matching. 

We take several measures to ensure a comprehensive and fair comparison. 
First, We update BYOC's visual backbone from ResNet-5 to ResNet-18. This results in a stronger and fairer baseline which we refer to as BYOC$^\dagger$.
Second, SuperGlue and LoFTR are both fully supervised image-based approaches which use transformers for feature matching. 
Hence, it was unclear how to adapt their matching algorithm to use depth information. Instead of adapting their matching, we use GART to re-rank their proposed matches and only retain the top set of matches. This resulted in large performance improvements as seen in ~\cref{tab:scannet_pairwise_corr}

\begin{table}[t]
\caption{\textbf{Wide-Baseline Correspondence Estimation on ScanNet.} 
SyncMatch extracts accurate wide-baseline correspondences; performing on-par with supervised methods. 
Our proposed GART uses estimated alignment to sample more accurate correspondences regardless of the underlying feature descriptor. 
}
\renewcommand{\arraystretch}{1.04}
\setlength\tabcolsep{4pt}
\renewcommand{\aboverulesep}{0pt}
\renewcommand{\belowrulesep}{0pt}
\centering
\resizebox{\linewidth}{!}{%
\centering
\begin{tabular}{l ccc ccc }
\toprule
& \multicolumn{3}{c}{ 3D Corres.}  
& \multicolumn{3}{c}{ 2D Corres.} 
\\
\cmidrule(lr){2-4} 
\cmidrule(lr){5-7}
Method 
& 1cm &  5cm & 10cm 
& 1px &  2px & 5px  
\\
\hline
\multicolumn{5}{l}{\textbf{\textit{Unsupervised Features with Heuristic Matching}}} \\
RootSIFT \cite{arandjelovic2012three}  & 
 7.1 & 29.2 & 35.1 & 2.8 & 8.6 & 22.9 \\
RootSIFT \cite{arandjelovic2012three} + GART &
14.1 & 72.4 & 84.3 &  3.8 & 12.8 & 42.8 \\
SuperPoint~\cite{detone2018superpoint}  & 
 7.5 & 41.4 & 51.3 & 2.5 & 8.6 & 29.5 \\
SuperPoint~\cite{detone2018superpoint}  + GART & 
16.8 & 73.7 & 84.3 & 4.7 & 15.5 & 47.9  \\
BYOC$^{\dagger}$~\cite{elbanani2021byoc}   &
12.8 & 53.3 & 63.0 &  4.5 & 14.6 & 41.9 \\
BYOC$^{\dagger}$~\cite{elbanani2021byoc} + GART &
22.8 & 73.1 & 81.4 &  6.0 & 19.6 & 54.0 \\
SyncMatch (\textbf{Ours})  &  
13.1 & 55.1 & 65.4 & 4.6 & 15.3 & 43.9 \\
SyncMatch (\textbf{Ours}) + GART &  
\textbf{26.8} & 76.5 & 84.4 &  \textbf{7.5} & \textbf{23.5} & \textbf{59.7} \\
\midrule
\multicolumn{5}{l}{\textbf{\textit{Supervised Features with Trained Matching}}}  \\
SuperGlue~\cite{sarlin2020superglue}  &  
8.7 & 62.4 & 78.7 & 2.5 & 9.0 & 36.9 \\
SuperGlue~\cite{sarlin2020superglue} + GART &  
13.8 & 74.8 & 87.7 & 3.3 & 11.7 & 44.4 \\
LoFTR~\cite{sun2021loftr}  &
16.0 & 72.2	& 84.6 & 5.6 & 18.5 & 55.5 \\
LoFTR~\cite{sun2021loftr} + GART &
21.4 & \textbf{80.8} & \textbf{90.2} &  6.5 & 21.2 & 59.0 \\
\bottomrule
\end{tabular}%
}
\label{tab:scannet_pairwise_corr}

\end{table}

\textbf{How does heuristic matching perform?}
We find that well-tuned matching allows hand-crafted features to achieve a strong performance against learned features, as observed by Efe~\etal~\cite{efe2021effect}. Nevertheless, self-supervised feature descriptors still retain a performance advantage. Furthermore, our proposed approach outperforms both hand-crafted and self-supervised descriptors regardless of whether depth is used at test time for refinement.

\nsparagraph{Can self-supervision replace correspondence supervision?}
While our approach outperforms classic and self-supervised approaches, it still underperforms the strongest supervised approaches. 
This is expected since we use pseudocorrespondence labels from short sequences, whereas supervised approaches are trained with view pairs that were sampled the same way as the test pairs~\cite{sarlin2020superglue}. Nevertheless, we argue that our approach is still promising, as it can match SuperGlue supervised features and matching despite being self-supervised and using the ratio-test for matching.

\lsparagraph{Does geometry based refinement help?}
We find that our proposed geometry-based refinement improved the performance for all methods. Furthermore, the improvement is most pronounced for our method, which performs on-par with supervised methods when using depth, and outperforms them for some thresholds.

\subsection{Pointcloud Registration}

\begin{table}[!t]
\caption{\textbf{Pairwise Registration on ScanNet.} 
SyncMatch outperforms all approaches on narrow-baseline registration, while under performing some methods for wide-baseline registration. WP-RANSAC results in large performance gains for all methods across metrics.  
}

\label{tab:scannet_pairwise_reg}

\setlength\tabcolsep{4pt}
\renewcommand{\aboverulesep}{0pt}
\renewcommand{\belowrulesep}{0pt}
\centering
\resizebox{\linewidth}{!}{%
\begin{tabular}{l cc cc}
\toprule

& \multicolumn{2}{c}{Narrow Baseline}  
& \multicolumn{2}{c}{Wide Baseline} 
\\
\cmidrule(lr){2-3} 
\cmidrule(lr){4-5}

& AUC$_{5^\circ}$ & AUC$_{10\text{cm}}$ & AUC$_{5^\circ}$ & AUC$_{10\text{cm}}$ 
\\
\hline
\multicolumn{5}{l}{\textbf{\textit{Unsupervised Features with Heuristic Matching}}}  \\
RootSIFT~\cite{arandjelovic2012three} + RANSAC &
36.7 & 28.9 & 15.5 & 11.1 \\
SuperPoint~\cite{detone2018superpoint} + RANSAC &
50.3 & 39.8 & 24.9 & 17.0 \\
RootSIFT~\cite{arandjelovic2012three} + \textbf{Ours} & 
84.3 & 77.9 & 64.4 & 52.3 \\
SuperPoint~\cite{detone2018superpoint} + \textbf{Ours} &
83.8 & 77.0 & 61.7 & 49.2 \\ 
BYOC$^{\dagger}$~\cite{elbanani2021byoc} &
84.6 & 77.6 & 60.4 & 48.4 \\
SyncMatch \textbf{(Ours)} &
85.3 & 78.8 & 63.4 & 50.5 \\
\midrule
\multicolumn{5}{l}{\textbf{\textit{Supervised Features with Trained Matching}}}  \\
SuperGlue~\cite{sarlin2020superglue} + RANSAC & 
65.7 & 54.0 & 47.6 & 33.2 \\
LoFTR~\cite{sun2021loftr} + RANSAC &
75.0 & 64.6 & 57.2 & 41.8 \\
SuperGlue~\cite{sarlin2020superglue} + \textbf{Ours} &
82.3 & 75.0 & 66.0 & 51.2 \\
LoFTR~\cite{sun2021loftr} + \textbf{Ours}  &
84.5 & 78.1 & 70.5 & 56.2 \\
\bottomrule
\end{tabular}%
}

\end{table}

We next evaluate pairwise and multiview pointcloud registration performance.
We evaluate pairwise registration using view pairs extracted from ScanNet.
We also evaluate our model's ability to scale up to large sequences by reconstructing the challenging sequences in the ETH-3D dataset.

\paragraph{Pairwise Registration. }
We evaluate the approaches on pairwise registration for both narrow-baseline and wide-baseline view pairs, as shown in~\cref{tab:scannet_pairwise_reg}. 
We evaluate narrow baseline view pairs similar to BYOC and wide-baseline view pairs similar to SuperGlue. 
We report the area under the curve for pose errors with a threshold of $5^\circ$ and $10$cm for rotation and translation errors, respectively. 
For RootSIFT and SuperPoint, we compute correspondences using the ratio test, while SuperGlue and LoFTR provide us with matches. We use either Open3D's~\cite{open3d} RANSAC or our proposed WP-RANSAC for alignment. 
For SyncMatch and BYOC, we use the method's estimated transformation. 
We report numbers without depth refinement to avoid confounding the evaluation. 

Our approach outperforms all baselines for narrow-baseline registration but underperforms several in wide-baseline registration. 
This is surprising given our model's strong performance in wide-baseline correspondence estimation, but probably arises from the domain shift from the mostly narrow-baseline pairs used for training. 
Furthermore, we note that our proposed alignment algorithm greatly improves the performance of all baselines, especially RootSIFT. 
We observe this improvement from supervised models with trained correspondence estimators, suggesting that their predicted correspondences still contain significant structured error that benefit from robust estimators that can utilize the match confidence weights. 

\begin{figure}[t!]
\begin{center}
\includegraphics[width=\linewidth]{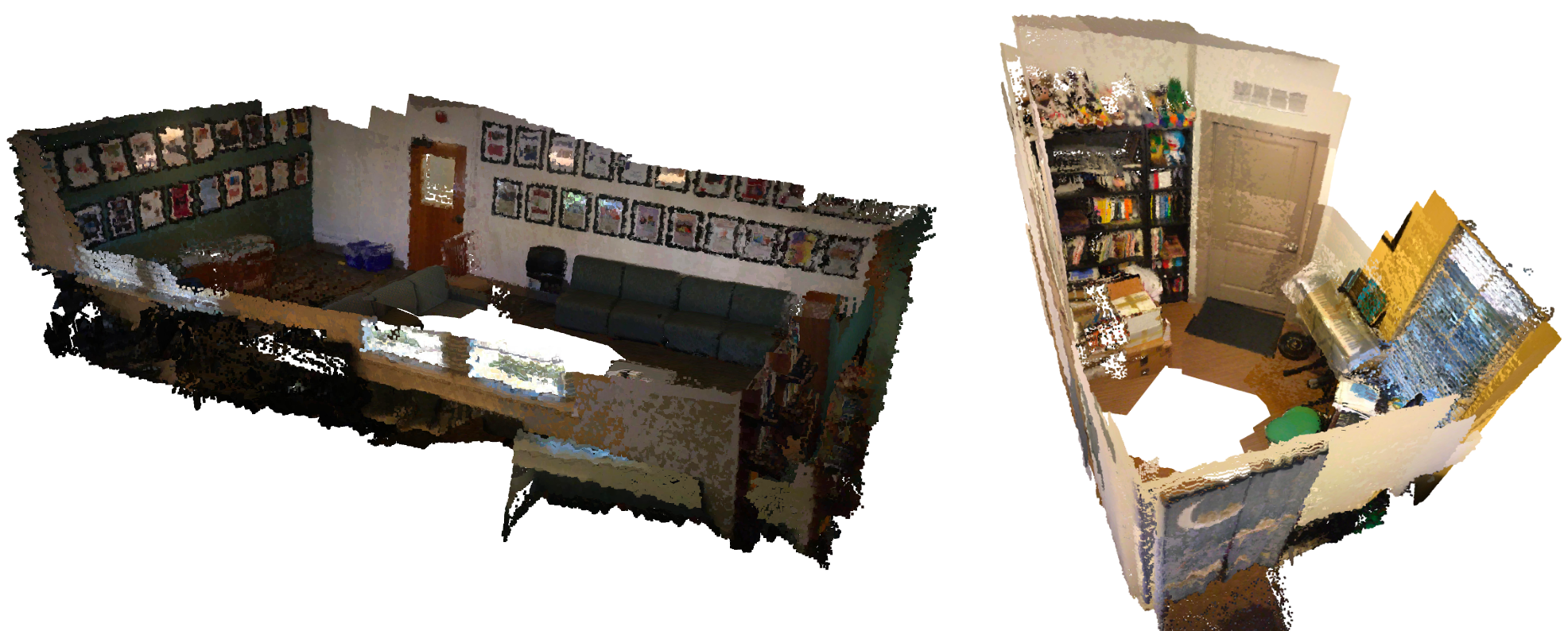}
\end{center}
\caption{\textbf{RGB-D Scene Reconstruction.} SyncMatch can scale at inference time to longer videos to reconstruct longer sequences. }
\label{fig:recon}
\end{figure}

\paragraph{Scaling up to longer sequences. }
Computing pairwise correspondence for $N$ frames scales as O$(N^2)$, which is problematic for longer videos. However, with minor reformulation, SyncMatch can be adapted to significantly reduce its run-time. Instead of considering all pairs, we only consider adjacent frames in the first step to give us an approximate camera trajectory from $N{-}1$ pairs. We use this trajectory to refine the correspondences and then consider all frames within a specific window $w$; \ie, only consider frames $(i, j)$ if $|i - j| < w$. We can then run the synchronization step with the confidence of all other pairs set to 0. This allows us to scale the model's run-time linearly with number of frames, instead of quadratically, which allows us to handle hundreds of frames. We visualize two reconstructions from ScanNet in~\cref{fig:recon}. 

We apply our adapted model to the challenging ETH3D dataset. 
ETH-3D sequences are challenging to traditional RGB-D reconstruction models, with BundleFusion failing on nearly 75\% of the sequences. 
SyncMatch can reconstruct 33/61 training scenes and 16/35 test scenes. 
This outperforms standard systems such as BundleFusion (14/61 and 7/35) and ORB-SLAM~\cite{murORB2} (25/61 and 16/35). 
Since such systems are often used to automatically generate annotations for RGB-D video datasets, SyncMatch's strong performance against them shows the promise of self-supervised approaches to use videos that currently are missing from large-scale RGB-D scene datasets.  
We emphasize that our model was not designed for full-scene reconstruction; this evaluation is only intended to showcase our model's performance against existing methods for automatically generating pose annotation for RGB-D videos. 

\subsection{Analysis}

We analyze our model through a series of experiments aimed at answering some key questions regarding its performance.
The analysis experiments are aimed at understanding the impact of the multiview setting, training data, as well as the impact of our model's components during both training and inference.

\begin{table}[t]
\caption{\textbf{Impact of number of training views.}
Training with more views improves correspondence estimation performance.
}
\renewcommand{\arraystretch}{1.04}
\setlength\tabcolsep{10pt}
\renewcommand{\aboverulesep}{0pt}
\renewcommand{\belowrulesep}{0pt}
\centering
\resizebox{\linewidth}{!}{%
\centering
\begin{tabular}{l ccc ccc }
\toprule
& \multicolumn{3}{c}{ 3D Corres.}  
& \multicolumn{3}{c}{ 2D Corres.} 
\\
\cmidrule(lr){2-4} 
\cmidrule(lr){5-7}
Num. of Views
& 1cm &  5cm & 10cm 
& 1px &  2px & 5px  
\\
\hline
2 & 24.7 & 73.9 & 81.6 &  6.2 & 19.8 & 54.0 \\
3 & 26.2 & 76.8 & 85.2 &  7.0 & 22.6 & 58.8 \\
4 & 26.8 & 75.4 & 83.5 &  7.5 & 23.3 & 59.2 \\
5 & 26.9 & 75.9 & 83.6 &  7.6 & 23.5 & 59.7 \\
6 & 26.8 & 76.5 & 84.4 &  7.5 & 23.5 & 59.7 \\
\bottomrule
\end{tabular}%
}
\label{tab:analysis_num_views}

\end{table}

\begin{table}[t]
\centering
\caption{\textbf{Ablation Experiments.}
While WP-RANSAC is crucial to model performance, the impact of other components depends on downstream tasks. 
}
\label{tab:ablations}

\setlength\tabcolsep{3pt}
\renewcommand{\aboverulesep}{0pt}
\renewcommand{\belowrulesep}{0pt}
\resizebox{\linewidth}{!}{%
\begin{tabular}{l cc cc cc}
\toprule
& \multicolumn{2}{c}{Ablation}  
& \multicolumn{2}{c}{PW Corr.}  
& \multicolumn{2}{c}{MV Reg.} 
\\
\cmidrule(lr){2-3} 
\cmidrule(lr){4-5} 
\cmidrule(lr){6-7}
& Train & Test & AUC$_{10cm}$ & AUC$_{10px}$ & AUC$_{5^\circ}$ & AUC$_{10cm}$
\\
\midrule
Full Model               & & & 
65.9 & 48.1 &  83.4 & 77.8 \\
\midrule
Naive Sync.    & \xmark & & 
65.7 & 48.9 & 82.8 & 76.2 \\
No Depth Refine    & \xmark & & 
66.0 & 49.0 & 82.9 & 76.3 \\
No Conf Threshold    & \xmark & & 
35.2 & 20.4 & 17.9 & 14.5 \\
No WP-RANSAC     & \xmark & & 
 1.2 &  0.2 & 1.1 & 3.3 \\
\midrule
Naive Sync.    &  & \xmark & 
65.9 & 48.1 & 82.6 & 76.9 \\
No Depth Refine    &  & \xmark & 
29.7 & 19.4, & 83.4 & 77.7 \\
No Conf. Threshold    &  & \xmark & 
65.9 & 48.1 & 76.9 & 70.9 \\
No WP-RANSAC    &  & \xmark & 
42.2 & 27.7 & 74.4 & 66.0 \\
\bottomrule
\end{tabular}%
}

\end{table}

\paragraph{What's the impact of the number of training views?}
We train our model with a variable number of training views to understand the impact of multiview training. 
We observe that increasing the number of training views results in progressively better performance. 
However, the performance gains saturate after 5 views. This could be explained how ScanNet videos were captured: often, the camera is moving laterally and after 5 frames (3 seconds), there is often no overlap. Our results suggest that using more views for training will not help until enough frames are used to provide loop closure for the model to learn from.

\lsparagraph{How do the different components contribute to the model performance?}
We analyze the impact of various model components through a series of ablations that are applied either during training or testing. 
We report the performance for both pairwise correspondence estimation (3D and 2D correspondence error) as well as multiview registration with 6 views (rotation and translation error) in~\cref{tab:ablations}. 
Similar to prior work~\cite{elbanani2021unsupervisedrr}, we observe that ablations that make it harder to register the pointcloud can boost correspondence performance when ablated during training; \eg, using naive synchronization based only on adjacent views or training without depth refinement. However, replacing WP-RANSAC with WP prevents the model from learning due to inaccurate registration early during training. We also observe that almost all test-time ablations result in worse performance. One surprising exception is removing depth refinement which greatly reduces wide-baseline correspondence estimation accuracy, while not impacting multiview registration. This could be explained by the performance saturating for narrow-baseline registration such that depth refinement is not needed there. 

\vspace{-0.8cm}
\msparagraph{Can we learn from more challenging sequences?}
While ScanNet offers a large set of videos, these videos are carefully captured to ensure successful downstream 3D reconstruction. 
We investigate whether our approach can be trained on more challenging videos, such as ETH-3D. ETH-3D has fewer and more erratic videos, forcing us to reduce the view stride during training. However, our model can still learn without supervision on ETH-3D videos and the features can be used for wide-baseline correspondence estimation as shown in~\cref{tab:analysis_eth}. For comparison, we train a on a subset of ScanNet to match number of instances and view spacing. Both models achieve similar performance. This suggests that our approach could scale to challenging videos beyond carefully captured ScanNet sequences.

\begin{table}[t]
\caption{\textbf{Training on ETH-3D data.}
Our model is capable of learning for the more challenging videos in ETH-3D. 
}
\renewcommand{\arraystretch}{1.04}
\setlength\tabcolsep{8pt}
\renewcommand{\aboverulesep}{0pt}
\renewcommand{\belowrulesep}{0pt}
\centering
\resizebox{\linewidth}{!}{%
\centering
\begin{tabular}{l ccc ccc }
\toprule
& \multicolumn{3}{c}{ 3D Corres.}  
& \multicolumn{3}{c}{ 2D Corres.} 
\\
\cmidrule(lr){2-4} 
\cmidrule(lr){5-7}
Training Set
& 1cm &  5cm & 10cm 
& 1px &  2px & 5px  
\\
\hline
ETH-3D          & 17.4 & 66.4 & 74.5 & 4.8 & 16.1 & 47.3 \\
ScanNet Mini    & 18.5 & 67.0 & 75.6 & 4.8 & 16.2 & 47.6 \\
\bottomrule
\end{tabular}%
}
\label{tab:analysis_eth}

\end{table}

\section{Conclusion}
\label{sec:conclusion}
We present SyncMatch: a self-supervised approach to learning correspondence estimation that relies on multiview registration to learn from difficult view pairs. 
Our core insight is that multiview registration allows us to leverage the consistency within short video sequences to obtain difficult view pairs for learning.
To this end, we propose a series of components that integrate self-supervised correspondence estimation with multiview registration in a single end-to-end differentiable pipeline. Our approach follows the conventional registration pipeline with several technical contributions that improve registration and correspondence estimation performance while maintaining differentiability. 

Our goal is not to beat supervised methods, but rather to show that supervision might not be needed. 
While standard 3D reconstruction systems like COLMAP and BundleFusion provide us with good reconstructions, learned approaches trained with their outputs are starting to rival their accuracy.
This raises an interesting question: how can we exceed standard pipelines when our training data is limited by what they can handle and our supervision is limited by their error?  
Through proposing self-supervised pipelines that learn directly from videos, we take a step towards enabling the development of approaches that go beyond conventional setups and scale to more uncurated data, allowing us to both achieve better 3D reconstruction and tackle difficult challenges like dynamic scene reconstruction.

{
\small
\vspace{0.2cm}
\noindent 
\textbf{Acknowledgments}
We thank Karan Desai, Mahmoud Azab, David Fouhey, Richard Higgins, Daniel Geng, and Menna El Banani for feedback and edits to early drafts of this work.
}


\clearpage
\onecolumn 
\begin{center}
\Large \bf
Supplementary Material: \\
Self-supervised Correspondence Estimation via Multiview Registration
\end{center}

\appendix
\section{Implementation Details}
\label{app:impl}
We include a subset of our source code in the supplemental material. 
The submitted version was taken from a larger code base and edited to improve clarity through comments and remove any identifying information.
Our approach is implemented in PyTorch~\cite{paszkepytorch}, but we make heavy use of PyKeOps~\cite{PyKeOps} and FAISS~\cite{FAISS} for fast CUDA implementations of kNN, as well as PyTorch3D~\cite{pytorch3d} and Open3D~\cite{open3d} for 3D transformations and alignment. 
We discuss several key design choices below and refer the reader to our code submission for the exact details. 
For more details on SE(3) Transformation Synchronization, we refer the reader to~\cref{app:synchronization}. 

\lsparagraph{Feature Extraction. }
We use a modified ResNet-18~\cite{he2016deep} as our feature extractor.
Since the ResNet architecture was designed for image classification, it performs aggressive downsampling to reduce the spatial dimension of the feature grid to allow the network to be more light-weight while increasing the receptive field for each pixel. However, our application would benefit from maintaining a high resolution to allow for accurate matching. As a result, we modify ResNet to remove most of the down-sampling, only down-sampling by a factor of $2{\times}$ twice within the network. During training, we down-sample the input to a dimension of $240{\times}320$. We find that this allows us to increase the speed of training, without impacting the test-time performance on images of resolution $480{\times}640$.
We set the output feature dimension to 128. 
While previous work on self-supervised learning was restricted to small feature dimensions due to a slow kNN implementation~\cite{elbanani2021byoc,elbanani2021unsupervisedrr}, we use the faster kNN implementations provided by PyKeOps and FAISS. 

\lsparagraph{Correspondence Estimation. }
We use the kNN functions provided by FAISS~\cite{FAISS} and PyKeOps~\cite{PyKeOps} in our implementation. While FAISS provides a faster kNN implementation, PyKeOps provides more flexibility in the distance function that can be used. As a result, we use FAISS for the initial feature-based correspondence estimation, and use PyKeOps for finding kNN based on both features and geometry for the geometry-aware ratio test. In all cases, we filter the correspondence and only keep the top 500 correspondences.

\lsparagraph{Confidence Threshold. }
We only apply confidence thresholding to non-adjacent frames; \ie, $|i - j| > 1$. We do this as we find that some adjacent pairs can still have a low pair-wise confidence despite having large overlap. 
Through excluding adjacent pairs from thresholding, we can guarantee that synchronization is possible for all sequences. We set the confidence threshold to $\gamma=0.4$ which allows us to ensure large overlap as shown in Figure 3 in the main paper. 

\lsparagraph{Refinement. }
Given the synchronized views, we resample correspondences based on both feature similarity and spatial proximity of points. This allows us to sample better correspondences as shown in Figure 4 of the main paper. We set the weighting between feature and spatial distance to $\alpha=10.0$ based on preliminary experiments.

\section{Qualitative Results}
\label{app:qualitative}
We include additional qualitative results to provide a better sense of our model's performance. We also clarify some of the color schemes used throughout the paper. 

\lsparagraph{Correspondence color.}
We color-code our correspondence using their 3D error. Specifically, correspondences with an error of less than 5 cm were plotted in dark green, errors between 5 cm and 10 cm were plotted in yellowish green, errors between 10 cm and 15 cm were plotted in orange, and errors larger than 15 cm were plotted in red. 

\lsparagraph{Correspondence Estimation. }
We provide additional qualitative examples of correspondence estimation results in~\cref{fig:app_corr}. 
We find that for easy cases that involve the camera panning or zooming, all approaches perform fairly well (rows 1-3). 
Meanwhile, cases with large camera rotation can be challenging to all models, with different models failing for different cases. We find that our model can overcome those challenges in some cases where some prior approaches have limited performance (rows 4-9). 
In cases with repeated textures, our model can inaccurately predict a consistent set of correspondences that are accurate, as shown in row 10 of~\cref{fig:app_corr}. 
We find that LoFTR can succeed in such a case, likely due to its use of cross-attention, which is noted by the authors of both LoFTR and SuperGlue. Future iterations of self-supervised correspondence estimation should explore the incorporation of attention modules and integrating it with geometric-aware matching.  
Finally, we observe that some cases are challenging to all models, especially when there is a very large camera motion such as looking at the same object from opposing sides (row 11) or when there is limited overlap and plain textures (row 12).

\lsparagraph{Correspondence Refinement. }
We provide qualitative examples of estimated correspondences before and after refinement in~\cref{fig:app_refine}.
In many cases, the initial feature-based correspondences are already fairly accurate. In those cases, we find that refinement results in the correspondences being more spread out and increasing in accuracy. 
More interesting cases involve a very noisy initial set that can be refined into a dense, accurate set of correspondences. This can be seen clearly in rows 5-7 in~\cref{fig:app_refine}. 
Finally, in some difficult cases, our initial estimation is extremely noisy, and our model is unable to recover from that.

\section{Camera Synchronization}
\label{app:synchronization}
\noindent Here we explain the Camera Synchronization algorithm (Section 3.4) in a bit more detail.

\lsparagraph{Notation for SE(3) matrices.}
Recall that for pairs of frame $i<j$,
\begin{equation*}
\mathbf{T}_{i,j} = \argmin_{\mathbf{T} \in \text{SE}(3)} \sum_{\text{inliers }(p, q, w) \in \mathcal{C}_{i,j}} w||\mathbf{x}_{q} - \mathbf{T}(\mathbf{x}_p)||_2^2
\end{equation*}
is our estimate for the relative transformation from camera $i$ to camera $j$. We can write $\mathbf{T}_{i,j}$ as a 4x4 matrix consisting of a rotation and translation,
\[
\mathbf{T}_{i,j}=\left[
\begin{array}{c|c}
R & 0 \\
\hline
t & 1
\end{array}
\right], \qquad R\in\text{SO}(3),\ T\in \mathbb{R}^3.
\]
$\mathbf{T}_{i,j}$ acts on points $\mathbf{x}=(x_1,x_2,x_3,1)$ in homogeneous form by right multiplication
\[
\mathbf{T}_{i,j}(\mathbf{x}) = (x_1,x_2,x_3,1) \times \mathbf{T}_{i,j}.
\]

\lsparagraph{Confidence-weighted transformations.}
Recall that $c_{i,j}$ is a confidence value attached to $\mathbf{T}_{i,j}$ for $i<j$. Let $\mathbf{S}_+\subset \mathbf{R}^{4\times 4}$ denote the set of $4\times 4$ matrices with the form:
\[
\alpha
\left[
\begin{array}{c|c}
R & 0 \\
\hline
t & 1
\end{array}
\right], 
\qquad \alpha\ge 0, R\in\mathbb{R}^{3\times 3},\ T\in \mathbb{R}^3.
\]
Elements of $\mathbf{S}_+$ can be projected onto SE$(3)$ by dividing by $\alpha$, and then using SVD to project $R$ onto SO$(3)$. 

Note that $\mathbb{S}_+$ is closed in the sense that if $A,B \in \mathbf{S}_+$ and $\alpha\ge0$, then $A+B$, $A\times B$ and $\alpha A$ are all in $\mathbf{S}_+$ too.

\lsparagraph{Confidences as jump probabilities} We will make two simpifying assumptions. First, we will assume that the $c_{i,j}$ have been scaled so that the rows sum to one: for all $i$, $\sum_j c_{i,j} =1$. Second, we assume that for each $i$, $c_{i,i+1}>0$. With these assumptions in place, $C=[c_{i,j}]$ is the stochastic matrix for an $N$-state Markov chain $(X_t)$,
\[
c_{i,j} = \mathbb{P} [ X_{t+1}= j \mid X_t=i], \quad t=0,1,2,\dots. 
\]
The Markov chain is \cite{LevinPeresWilmer2006}:
\begin{itemize}
\item lazy: $\mathbb{P}[X_1=i\mid X_0=i]=c_{i,i}=1/2$ as $c_{i,i}=\sum_{j\not=i} c_{i,j}$,
\item connected: for all $i,j$,  for some $t$ sufficiently large $\mathbb{P}[X_t=j\mid X_0=i]=(C^t)_{i,j}>0$, and
\item time-reversible: $\pi_i C_{i,j} = \pi_j C_{j,i}$ for all $i,j$ with $\pi\in[0,1]^N$ the equiilibrium distribution.
\end{itemize}
By the Perron–Frobenius theorem, and the laziness property, the eigenvalues of $C$ can be written as 
\begin{equation}\label{eq:sup-sync-lambda}
1=\lambda_1 > \lambda_2 \ge ... \lambda_N \ge 0.
\end{equation}
The spectral gap $1-\lambda_2>0$ so convergence to equilibrium is exponential,
\[
\mathbb{P}[X_t=j\mid X_0=i]=(C^t)_{i,j} = \pi_j + \text{O}(\lambda_2^t).
\]

\lsparagraph{The pairwise-transformations matrix}
In Section 3.4, equation (5), we define a $4N\times 4N$ matrix $\mathbf{A}$,
\begin{equation*}
    \mathbf{A}  = \begin{bmatrix*}[c] 
    c_{1,1} \mathbf{I}_4 & c_{1, 2}\mathbf{T}_{1, 2} & \cdots & c_{1, N}\mathbf{T}_{1, N}\\
    c_{2, 1}\mathbf{T}_{2, 1} & c_{2} \mathbf{I}_4 & \cdots & c_{2, N}\mathbf{T}_{2, N}\\
    \vdots & \ddots &  & \vdots\\
    \vdots &  & \ddots & \vdots\\
    c_{N, 1}\mathbf{T}_{N, 1} &  c_{N, 2}\mathbf{T}_{N, 2} & \cdots & c_{N,N} \mathbf{I}_4 \\
\end{bmatrix*}\in \mathbf{S}_+^{N\times N}\subset \mathbb{R}^{4N\times 4N},
\end{equation*}
consisting of an $N\times N$ grid of elements of $\mathbf{S}_+$.
To motivate the definition of $\mathbf{A}$, we can interpret it as generating a random walk on the set $\{1,2,\dots,N\}\times SE(3)$,
\[
\mathbb{P}[(X_{t+1},Y_{t+1})=( j,Y_t \times \mathbf{T}_{i,j}) \mid X_t=i ] = c_{i,j}, \qquad {t=0,1,\dots}
\]
The expected value in $\mathbf{S}_+$ of the $Y$-component of the walk is a weighted sum of the products of pairwise transformations. Pairwise transformations with greater confidence contribute more strongly to the sum.  

The solution to the synchronization problem is related to the eigenvectors of $\mathbf{A}$.
{\it If} there is a global collection of cameras $(\mathbf{T}_i)$ such that $\mathbf{T}_{i,j}=\mathbf{T}_i^{-1} \mathbf{T}_j$, then $\mathbf{A}$ will have four independent eigenvectors with eigenvalue one, i.e.
\[
[\mathbf{T}_1 \dots \mathbf{T}_N] \times (\mathbf{A} -\mathbf{I}_{4N}) = \mathbf{0}, \qquad [\mathbf{T}_1 \dots \mathbf{T}_N] \in \text{SE}(3)^N \subset \mathbb{R}^{4\times 4N}.
\]
All other eigenvalues $\lambda$ will satisfy $|\lambda| \le \lambda_2$ (c.f. inequality \eqref{eq:sup-sync-lambda} for the eigenvalues of Markov chain $X_t$, and properties of matrix determinants). As integer $k\to\infty$, each of the $N$ rows of $\mathbf{A}^k$ will converge to a globally consistent set of cameras. The different rows will yield essentially the same solution, but differing by an SE$(3)$ transformation of the global coordinates.

More generally, if no such perfect solution exists, then we want to find
\begin{align*}
\argmin_{\{\mathbf{T}_i\in\text{SE}(3):1\le i \le N\}}& \| [\mathbf{T}_1 \dots \mathbf{T}_N] \times (\mathbf{A} -\mathbf{I}_{4N}) \|_F^2\\
&=
\argmin_{\{\mathbf{T}_i\in\text{SE}(3):1\le i \le N\}} 
\sum_j \left\| \mathbf{T}_j - \sum_i c_{i,j} \mathbf{T}_i \mathbf{T}_{i,j} \right\|_F^2\\
&=
\argmin_{\{\mathbf{T}_i\in\text{SE}(3):1\le i \le N\}} 
\sum_{i,j} c_{i,j} \left\| \mathbf{T}_j - \mathbf{T}_i \mathbf{T}_{i,j} \right\|_F^2.
\end{align*}

The solution in \cite{gojcic2020learning} involves calculating an eigen decomposition directly. Let $\mathbf{A}^\text{rot}$ denote the $3N\times3N$ matrix obtained from $\mathbf{A}$ by taking the top $3\times3$ elements from each sub-block of $A$. In the notation of \cite[supplementary Sec. 2]{gojcic2020learning}, our $\mathbf{A}^\text{rot}$ is equal to their ``$L/2+D$''. Each of the $3N$ eigenvectors of $\mathbf{A}^\text{rot}$ (suitably padded with zeros to increase their length from $3N$ to $4N$, e.g.
\[
[x_1,x_2,x_3,\dots,x_{3N-2},x_{3N-1},x_{3N}] \rightarrow [x_1, x_2, x_3, 0, \dots,x_{3N-2},x_{3N-1},x_{3N}, 0] ),
\]
becomes an eigenvectors of $\mathbf{A}$. Three of these eigenvectors with largest eigenvalues, projected onto $SO(3)$, solve \cite[Eq.~5]{gojcic2020learning},
\[
\argmin_{\{R_i \in SO(3):1\le i \le N\}} \sum_{i,j} c_{i,j} \| R_j- R_i \times (\mathbf{T}_{i,j})_{1:3,1:3} \|_F^2
\]
Rather than computing the eigendecomposition of $\mathbf{A}$ directly, we instead use power-iteration. Raising $\mathbf{A}$ to large powers filters out the effect of the smaller eigenvalues. To do this efficiently, starting from $\mathbf{A}$, we repeatedly takes squares to calculate $\mathbf{A}^2$, then $\mathbf{A}^4$, and so on until $\mathbf{A}^{2^t}$. Each element in $\mathbf{A}^{2^t}$ is then projected into SE$(3)$ using SVD as described above; call the resulting matrix $\mathbb{A}$.
$\mathbb{A}$ is composed of $N$ `rows', each with shape $4\times 4N$; each of these rows is n approximate solution to the synchronization problem. The difference between the rows is that each row is centered around a different camera. We choose $t=O(\log N)$ so $2^t>N$; the the number of FLOPs needed to calculate $\mathbb{A}$ is thus O$(N^3 \log N)$. In practice, the time spent on synchronization is small compared to feature extraction and matching, as synchronization is independent of the resolution of the images. For very larger $N$, a database of key-frames could be used to reduce the size of $N$.

\begin{figure}[t!]
\begin{center}
\includegraphics[width=\linewidth]{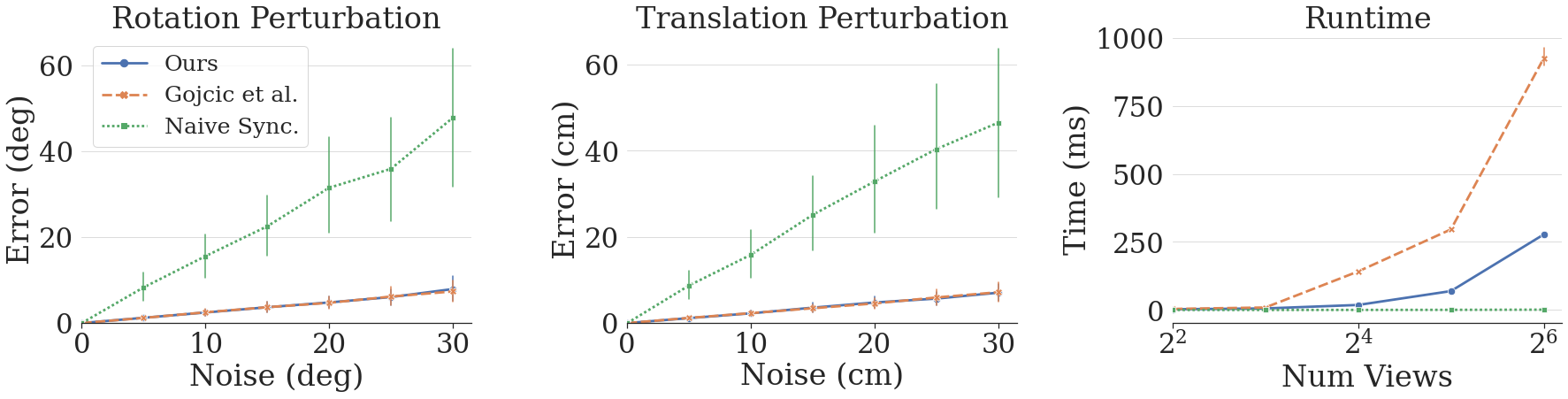}
\end{center}
\caption{\textbf{Synchronization Benchmark.} Our approach achieves the same error as Gojcic~\etal~\cite{gojcic2020learning}, while being faster and more numerically stable.}
\label{fig:benchmark_sync}
\end{figure}

\paragraph{Numerical stability.} Empirically, we find that training using synchronizations extracted  from $\mathbf{A}^{2^t}$ was stable. Using an eigensolver to implement the method of \cite{gojcic2020learning} led to exploding gradients. The derivative with respect to a set of eigenvectors is unstable when the eigenvalues a clustered together, as is normally the case with the largest eigenvalues of $\mathbf{A}^\text{rot}$; when the pairwise rotations are compatible, the largest eigenvalues will be approximately equal. 

\paragraph{Performance.}
We compare our synchronization approach to naive synchronization which aggregates the transformations using adjacent views and the eigendecomposition approach proposed by Gojcic~\etal~\cite{gojcic2020learning}. We compare the three algorithms on their ability to handle rotation and translation perturbation in the pairwise estimates as well as their runtime. As seen in \cref{fig:benchmark_sync}, our approach achieves the same performance as the eigendecomposition approach while being faster. Both approaches greatly naive synchronization since they are able to use information from all pairs. Furthermore, since our approach only relies on power iteration, it does not suffer from the numerical instability in the backward gradient discussed above.


\section{Ethical Considerations and Societal Implications}
\label{app:ethics}
Our work presents a method for self-supervised learning of correspondence estimation from RGB-D video. 
Our main contribution is to demonstrate how multiview registration could be used to learn better features from RGB-D video that perform on-par with supervised learning approaches. 
Advancements in this area can enable more powerful feature learners, which can improve the overall performance of larger frameworks that use it such as SLAM or structure-for-motion.
Our technical contributions pertain to allowing models to learn from a different type of data, as a result, our societal impact is mediated by the kind of data that we currently train on as well as the data that we can train on in the future. 

In this work, we evaluated our approach on ScanNet~\cite{dai2017scannet}. 
This is a large scale dataset of indoor scenes that contains over 1500 RGB-D sequences taken at more than 700 locations.
The data was collected by 20 volunteers across several countries with most sequences captured in the United States of America and Germany. 
Each volunteer used a specialized capture setup to record video sequences in private locations to which they had access. With very few exceptions, the data set is made up of empty rooms.
One salient issue with ScanNet is that most of the locations come from areas with a relatively high household income: houses, offices, and university housing in major western cities. 
This is problematic given that prior work has shown that computer vision models trained on data from countries with high household income generalize poorly to images coming from countries with a lower mean household income~\cite{de2019does}. 
Since our models were all trained on this dataset, we would expect them to generalize poorly to images from outdoor scenes or indoor scenes coming from different geographic areas or demographics. 
We note that while this could be alleviated by training our model on datasets coming from other countries, we are unaware of any large RGB-D video datasets that would meet such criteria. 

While our approach was trained on ScanNet, we developed a self-supervised approach since we hope that it can scale to large-scale data. 
Given the increasing prevalence of RGB-D sensors on phones, we can expect that people will start uploading videos to the web.
Such videos will not be careful scans of scenes as the ones captured for ScanNet, and hence will not be well-suited for 3D reconstruction algorithms. 
Our goal is to build systems that could easily leverage this data by being self-supervised.
However, such web-data will introduce other complications. Below, we reflect on two ethical considerations: structural biases in representation and privacy. 

\lsparagraph{Structural Bias in Representation. }
While the RGB-D cameras are becoming more available to consumers, their relatively high cost means that that they will be adopted more by wealthier demographics. 
Hence, while we expect that the uploaded videos will be more diverse than ScanNet, the high cost of the devices means that they will likely be used primarily by wealthier individuals resulting in a bias in representation both across and within different countries. 
As a result, anyone considering using this technology for learning features should be aware of such limitations, which could potentially be alleviated by a more thoughtful collection process. 

\lsparagraph{Privacy Concerns. }
People often upload videos that include other individuals who may have not consented for videos to be uploaded or even captured with them. 
Such videos will inevitably include such individuals. This means that to scale our approach to such data, we need robust filtering mechanisms that can detect such shots and exclude them from training. 
This is a common challenge for any self-supervised method that hopes to scale to web data.

\begin{figure*}[t!]
\begin{center}
\includegraphics[width=\textwidth]{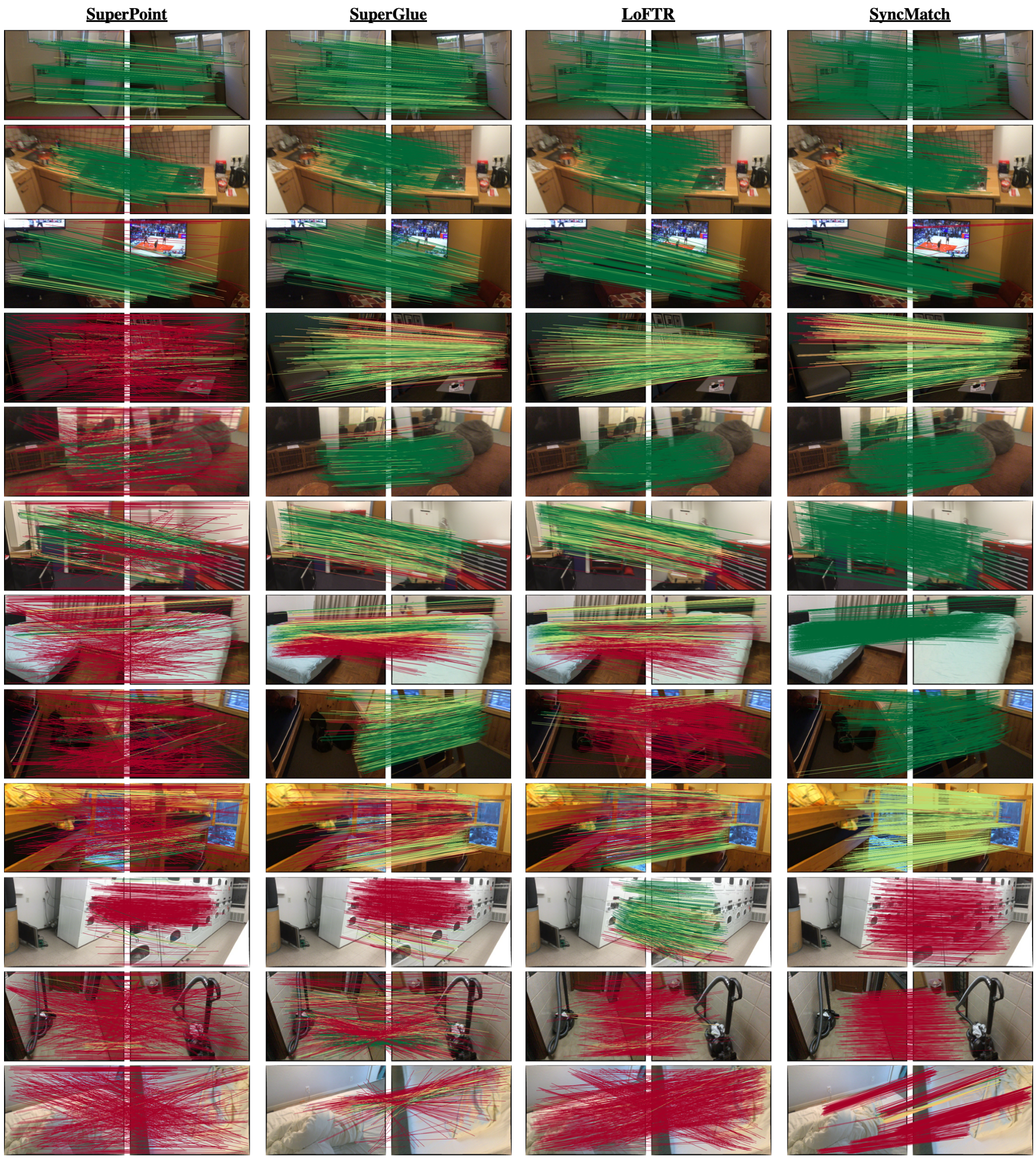}
\end{center}

\caption{\textbf{Correspondence Estimation.} 
We present correspondence results on a variety of situations. 
The top three rows show a variety of positive examples where all models performs well. 
The following six column present present cases where our model succeeds and other models perform poorly. Those are typically cases with large camera motion where our model is capable of use the geometric information and learned features to predict accurate correspondences. 
Finally, we report some challenging cases for our model: repetitive textures (row 10), very large camera motion (row 11), limited overlap and plain textures (row 12). While such cases are often challenging for all models, prior approaches like SuperGlue and LoFTR can sometimes produce good correspondence for some instances. 
}
\label{fig:app_corr}
\end{figure*}
\begin{figure*}[t!]
\begin{center}
\includegraphics[width=\textwidth]{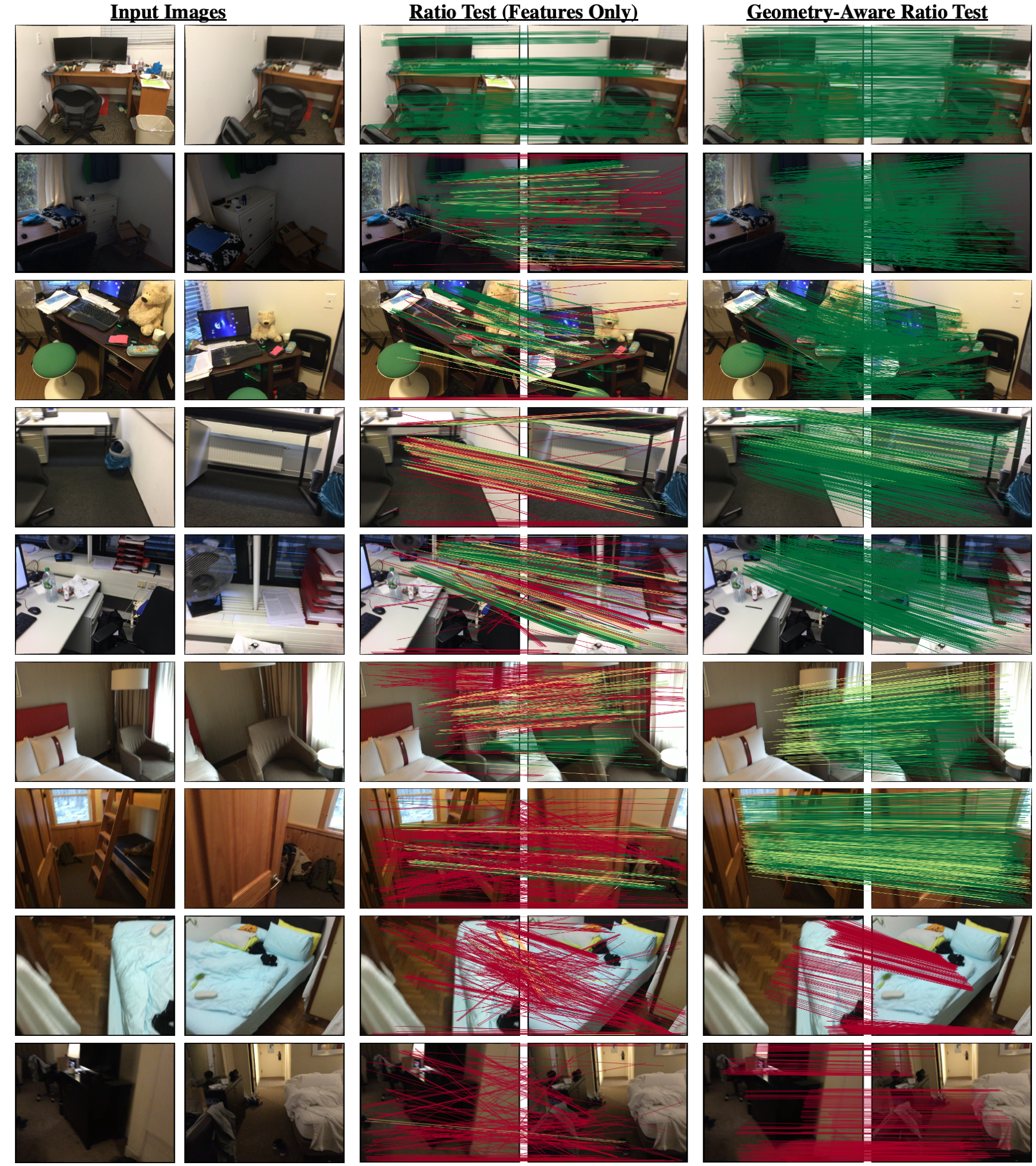}
\end{center}

\caption{\textbf{Correspondence Refinement.}
We show the impact of correspondence refinement using depth. We present several modes of performance. 
The top three rows present cases where the feature-based correspondences were already accurate and incorporating geometry simply improved the accuracy. 
The following three four rows cases where refinement had a large impact on correspondence quality by leveraging a small subset of accurate correspondences to align the scenes and then sample a more accurate set of correspondences. 
Finally, we observe failure cases where the initial set is so noisy that the model cannot generate a good transformation estimate rendering refinement useless. 
}
\label{fig:app_refine}
\end{figure*}

\clearpage

\clearpage

\twocolumn 

{\small
\bibliographystyle{ieee_fullname}
\bibliography{egbib}

\begin{thebibliography}{10}\itemsep=-1pt

\bibitem{arandjelovic2012three}
Relja Arandjelovi{\'c} and Andrew Zisserman.
\newblock Three things everyone should know to improve object retrieval.
\newblock In {\em CVPR}, 2012.

\bibitem{arrigoni2014robust}
Federica Arrigoni, Luca Magri, Beatrice Rossi, Pasqualina Fragneto, and Andrea
  Fusiello.
\newblock Robust absolute rotation estimation via low-rank and sparse matrix
  decomposition.
\newblock In {\em 3DV}, 2014.

\bibitem{arrigoni2016spectral}
Federica Arrigoni, Beatrice Rossi, and Andrea Fusiello.
\newblock Spectral synchronization of multiple views in se (3).
\newblock {\em SIAM Journal on Imaging Sciences}, 2016.

\bibitem{arun1987least}
KS Arun, TS Huang, and SD Blostein.
\newblock Least square fitting of two 3-d point sets.
\newblock In {\em TPAMI}, 1987.

\bibitem{balntas2016learning}
Vassileios {Balntas}, Edgar {Riba}, Daniel {Ponsa}, and Krystian {Mikolajczyk}.
\newblock Learning local feature descriptors with triplets and shallow
  convolutional neural networks.
\newblock In {\em BMVC}, 2016.

\bibitem{bay2006surf}
Herbert Bay, Tinne Tuytelaars, and Luc Van~Gool.
\newblock Surf: Speeded up robust features.
\newblock In {\em ECCV}, 2006.

\bibitem{birdal2020measure}
Tolga Birdal, Michael Arbel, Umut Şimşekli, and Leonidas Guibas.
\newblock Synchronizing probability measures on rotations via optimal
  transport.
\newblock In {\em CVPR}, 2020.

\bibitem{brachmann2017dsac}
Eric Brachmann, Alexander Krull, Sebastian Nowozin, Jamie Shotton, Frank
  Michel, Stefan Gumhold, and Carsten Rother.
\newblock {DSAC: Differentiable RANSAC for Camera Localization}.
\newblock In {\em CVPR}, 2017.

\bibitem{brachmann2019neural}
Eric Brachmann and Carsten Rother.
\newblock Neural-guided ransac: Learning where to sample model hypotheses.
\newblock In {\em ICCV}, 2019.

\bibitem{carlone2015initialization}
Luca Carlone, Roberto Tron, Kostas Daniilidis, and Frank Dellaert.
\newblock {Initialization techniques for 3D SLAM: a survey on rotation
  estimation and its use in pose graph optimization}.
\newblock In {\em ICRA}, 2015.

\bibitem{chang2017matterport3d}
Angel Chang, Angela Dai, Thomas Funkhouser, Maciej Halber, Matthias Niebner,
  Manolis Savva, Shuran Song, Andy Zeng, and Yinda Zhang.
\newblock {Matterport3D: Learning from RGB-D Data in Indoor Environments}.
\newblock In {\em 3DV}, 2017.

\bibitem{PyKeOps}
Benjamin Charlier, Jean Feydy, Joan~Alexis Glaunès, François-David Collin,
  and Ghislain Durif.
\newblock Kernel operations on the gpu, with autodiff, without memory
  overflows.
\newblock {\em JMLR}, 2021.

\bibitem{choy2020deep}
Christopher Choy, Wei Dong, and Vladlen Koltun.
\newblock Deep global registration.
\newblock In {\em CVPR}, 2020.

\bibitem{choy2019fully}
Christopher Choy, Jaesik Park, and Vladlen Koltun.
\newblock Fully convolutional geometric features.
\newblock In {\em ICCV}, 2019.

\bibitem{dai2017scannet}
Angela Dai, Angel~X. Chang, Manolis Savva, Maciej Halber, Thomas Funkhouser,
  and Matthias Nie{\ss}ner.
\newblock {ScanNet: Richly-annotated 3D Reconstructions of Indoor Scenes}.
\newblock In {\em CVPR}, 2017.

\bibitem{dai2017bundlefusion}
Angela Dai, Matthias Nie{\ss}ner, Michael Zoll{\"o}fer, Shahram Izadi, and
  Christian Theobalt.
\newblock {BundleFusion: Real-time Globally Consistent 3D Reconstruction using
  On-the-fly Surface Re-integration}.
\newblock {\em ACM ToG}, 2017.

\bibitem{de2019does}
Terrance de Vries, Ishan Misra, Changhan Wang, and Laurens van~der Maaten.
\newblock Does object recognition work for everyone?
\newblock In {\em Proceedings of the IEEE/CVF Conference on Computer Vision and
  Pattern Recognition Workshops}, pages 52--59, 2019.

\bibitem{deng2019directreg}
Haowen Deng, Tolga Birdal, and Slobodan Ilic.
\newblock 3d local features for direct pairwise registration.
\newblock In {\em CVPR}, 2019.

\bibitem{detone2018superpoint}
Daniel DeTone, Tomasz Malisiewicz, and Andrew Rabinovich.
\newblock {SuperPoint: Self-supervised interest point detection and
  description}.
\newblock In {\em CVPR Workshops}, 2018.

\bibitem{efe2021effect}
Ufuk Efe, Kutalmis~Gokalp Ince, and A~Aydin Alatan.
\newblock Effect of parameter optimization on classical and learning-based
  image matching methods.
\newblock In {\em CVPR}, 2021.

\bibitem{elbanani2021unsupervisedrr}
Mohamed El~Banani, Luya Gao, and Justin Johnson.
\newblock {UnsupervisedR\&R: Unsupervised Point Cloud Registration via
  Differentiable Rendering}.
\newblock In {\em CVPR}, 2021.

\bibitem{elbanani2021byoc}
Mohamed El~Banani and Justin Johnson.
\newblock {Bootstrap Your Own Correspondences}.
\newblock In {\em ICCV}, 2021.

\bibitem{fischler1981ransac}
Martin~A. Fischler and Robert~C. Bolles.
\newblock Random sample consensus: A paradigm for model fitting with
  applications to image analysis and automated cartography.
\newblock {\em Commun. ACM}, 1981.

\bibitem{gojcic2020learning}
Zan Gojcic, Caifa Zhou, Jan~D. Wegner, Leonidas~J. Guibas, and Tolga Birdal.
\newblock Learning multiview 3d point cloud registration.
\newblock In {\em CVPR}, 2020.

\bibitem{gojcic2019perfect}
Zan Gojcic, Caifa Zhou, Jan~D Wegner, and Andreas Wieser.
\newblock The perfect match: 3d point cloud matching with smoothed densities.
\newblock In {\em CVPR}, 2019.

\bibitem{graham2015sparse3dcnn}
Benjamin Graham.
\newblock {Sparse 3D convolutional neural networks}.
\newblock In {\em BMVC}, 2015.

\bibitem{he2016deep}
Kaiming He, Xiangyu Zhang, Shaoqing Ren, and Jian Sun.
\newblock Deep residual learning for image recognition.
\newblock In {\em CVPR}, pages 770--778, 2016.

\bibitem{hertz2020pointgmm}
Amir Hertz, Rana Hanocka, Raja Giryes, and Daniel Cohen-Or.
\newblock {PointGMM: A Neural GMM Network for Point Clouds}.
\newblock In {\em CVPR}, 2020.

\bibitem{huang2021predator}
Shengyu Huang, Zan Gojcic, Mikhail Usvyatsov, and Konrad~Schindler
  Andreas~Wieser.
\newblock {PREDATOR: Registration of 3D Point Clouds with Low Overlap}.
\newblock In {\em CVPR}, 2021.

\bibitem{huang2017translation}
Xiangru Huang, Zhenxiao Liang, Chandrajit Bajaj, and Qixing Huang.
\newblock Translation synchronization via truncated least squares.
\newblock {\em NeurIPS}, 2017.

\bibitem{huang2019learning}
Xiangru Huang, Zhenxiao Liang, Xiaowei Zhou, Yao Xie, Leonidas~J Guibas, and
  Qixing Huang.
\newblock Learning transformation synchronization.
\newblock In {\em CVPR}, 2019.

\bibitem{huang2020featureregistration}
Xiaoshui Huang, Guofeng Mei, and Jian Zhang.
\newblock Feature-metric registration: A fast semi-supervised approach for
  robust point cloud registration without correspondences.
\newblock In {\em CVPR}, 2020.

\bibitem{huber2003fully}
Daniel~F Huber and Martial Hebert.
\newblock Fully automatic registration of multiple 3d data sets.
\newblock {\em Image and Vision Computing}, 2003.

\bibitem{jabri2020walk}
Allan Jabri, Andrew Owens, and Alexei~A Efros.
\newblock Space-time correspondence as a contrastive random walk.
\newblock {\em NeurIPS}, 2020.

\bibitem{FAISS}
Jeff Johnson, Matthijs Douze, and Hervé Jégou.
\newblock Billion-scale similarity search with gpus.
\newblock {\em IEEE Transactions on Big Data}, 2021.

\bibitem{kabsch1976solution}
Wolfgang Kabsch.
\newblock A solution for the best rotation to relate two sets of vectors.
\newblock {\em Acta Crystallographica Section A: Crystal Physics, Diffraction,
  Theoretical and General Crystallography}, 1976.

\bibitem{kingma2015adam}
Diederik Kingma and Jimmy Ba.
\newblock Adam: A method for stochastic optimization.
\newblock In {\em ICLR}, 2015.

\bibitem{kummerle2011g}
Rainer K{\"u}mmerle, Giorgio Grisetti, Hauke Strasdat, Kurt Konolige, and
  Wolfram Burgard.
\newblock {G2o: A general framework for graph optimization}.
\newblock In {\em ICRA}. IEEE, 2011.

\bibitem{laguna2019key}
Axel~Barroso {Laguna}, Edgar {Riba}, Daniel {Ponsa}, and Krystian
  {Mikolajczyk}.
\newblock {Key.Net}: Keypoint detection by handcrafted and learned cnn filters.
\newblock In {\em ICCV}, 2019.

\bibitem{lai2019sslvidcorrflow}
Zihang Lai and Weidi Xie.
\newblock Self-supervised learning for video correspondence flow.
\newblock In {\em BMVC}, 2019.

\bibitem{le2019sdrsac}
Huu~M Le, Thanh-Toan Do, Tuan Hoang, and Ngai-Man Cheung.
\newblock {SDRSAC: Semidefinite-based randomized approach for robust point
  cloud registration without correspondences}.
\newblock In {\em CVPR}, 2019.

\bibitem{lenc2016learning}
Karel {Lenc} and Andrea {Vedaldi}.
\newblock Learning covariant feature detectors.
\newblock In {\em ECCV}, 2016.

\bibitem{LevinPeresWilmer2006}
David~A. Levin, Yuval Peres, and Elizabeth~L. Wilmer.
\newblock {\em {Markov chains and mixing times}}.
\newblock American Mathematical Society, 2006.

\bibitem{MegaDepthLi18}
Zhengqi Li and Noah Snavely.
\newblock Megadepth: Learning single-view depth prediction from internet
  photos.
\newblock In {\em CVPR}, 2018.

\bibitem{loshchilov2018decoupled}
Ilya Loshchilov and Frank Hutter.
\newblock Decoupled weight decay regularization.
\newblock In {\em ICLR}, 2018.

\bibitem{lowe2004distinctive}
David~G Lowe.
\newblock Distinctive image features from scale-invariant keypoints.
\newblock {\em IJCV}, 2004.

\bibitem{lu2019deepvcp}
Weixin Lu, Guowei Wan, Yao Zhou, Xiangyu Fu, Pengfei Yuan, and Shiyu Song.
\newblock Deepvcp: An end-to-end deep neural network for point cloud
  registration.
\newblock In {\em ICCV}, 2019.

\bibitem{melekhov2021digging}
Iaroslav Melekhov, Zakaria Laskar, Xiaotian Li, Shuzhe Wang, and Juho Kannala.
\newblock Digging into self-supervised learning of feature descriptors.
\newblock In {\em 3DV}, 2021.

\bibitem{mikolajczyk2004scale}
Krystian Mikolajczyk and Cordelia Schmid.
\newblock Scale \& affine invariant interest point detectors.
\newblock {\em IJCV}, 2004.

\bibitem{mur2015orb}
Raul Mur-Artal, Jose Maria~Martinez Montiel, and Juan~D Tardos.
\newblock Orb-slam: a versatile and accurate monocular slam system.
\newblock {\em {IEEE Transactions on Robotics}}, 2015.

\bibitem{murORB2}
Ra\'ul Mur-Artal and Juan~D. Tard\'os.
\newblock {ORB-SLAM2}: an open-source {SLAM} system for monocular, stereo and
  {RGB-D} cameras.
\newblock {\em IEEE Transactions on Robotics}, 2017.

\bibitem{page1999pagerank}
Lawrence Page, Sergey Brin, Rajeev Motwani, and Terry Winograd.
\newblock The pagerank citation ranking: Bringing order to the web.
\newblock Technical report, Stanford InfoLab, 1999.

\bibitem{paszkepytorch}
Adam Paszke, Soumith Chintala, Ronan Collobert, Koray Kavukcuoglu, Clement
  Farabet, Samy Bengio, Iain Melvin, Jason Weston, and Johnny Mariethoz.
\newblock Pytorch: Tensors and dynamic neural networks in python with strong
  gpu acceleration, may 2017.

\bibitem{ranftl2018deepfundamental}
Rene Ranftl and Vladlen Koltun.
\newblock Deep fundamental matrix estimation.
\newblock In {\em ECCV}, 2018.

\bibitem{pytorch3d}
Nikhila Ravi, Jeremy Reizenstein, David Novotny, Taylor Gordon, Wan-Yen Lo,
  Justin Johnson, and Georgia Gkioxari.
\newblock Accelerating 3d deep learning with pytorch3d.
\newblock {\em arXiv}, 2020.

\bibitem{rocco2017convolutional}
Ignacio {Rocco}, Relja {Arandjelovic}, and Josef {Sivic}.
\newblock Convolutional neural network architecture for geometric matching.
\newblock In {\em CVPR}, 2017.

\bibitem{rocco2018neighbourhood}
Ignacio Rocco, Mircea Cimpoi, Relja Arandjelovi{\'c}, Akihiko Torii, Tomas
  Pajdla, and Josef Sivic.
\newblock Neighbourhood consensus networks.
\newblock In {\em NeurIPS}, 2018.

\bibitem{roessle2022end2end}
Barbara Roessle and Matthias Nie{\ss}ner.
\newblock End2end multi-view feature matching using differentiable pose
  optimization.
\newblock {\em arXiv preprint arXiv:2205.01694}, 2022.

\bibitem{salas2013slam++}
Renato~F Salas-Moreno, Richard~A Newcombe, Hauke Strasdat, Paul~HJ Kelly, and
  Andrew~J Davison.
\newblock Slam++: Simultaneous localisation and mapping at the level of
  objects.
\newblock In {\em CVPR}, 2013.

\bibitem{sarlin2020superglue}
Paul-Edouard Sarlin, Daniel DeTone, Tomasz Malisiewicz, and Andrew Rabinovich.
\newblock {SuperGlue: Learning feature matching with graph neural networks}.
\newblock In {\em CVPR}, 2020.

\bibitem{schmid1997local}
Cordelia Schmid and Roger Mohr.
\newblock Local grayvalue invariants for image retrieval.
\newblock {\em TPAMI}, 1997.

\bibitem{schonberger2016structure}
Johannes~L Schonberger and Jan-Michael Frahm.
\newblock Structure-from-motion revisited.
\newblock In {\em CVPR}, 2016.

\bibitem{schoeps2017cvpr}
Thomas Sch\"ops, Johannes~L. Sch\"onberger, Silvano Galliani, Torsten Sattler,
  Konrad Schindler, Marc Pollefeys, and Andreas Geiger.
\newblock {A Multi-View Stereo Benchmark with High-Resolution Images and
  Multi-Camera Videos}.
\newblock In {\em CVPR}, 2017.

\bibitem{snavely2006photo}
Noah Snavely, Steven~M Seitz, and Richard Szeliski.
\newblock Photo tourism: exploring photo collections in 3d.
\newblock In {\em ACM SIGGRAPH}, 2006.

\bibitem{sun2021loftr}
Jiaming Sun, Zehong Shen, Yuang Wang, Hujun Bao, and Xiaowei Zhou.
\newblock {LoFTR}: Detector-free local feature matching with transformers.
\newblock {\em CVPR}, 2021.

\bibitem{tian2017l2}
Yurun {Tian}, Bin {Fan}, and Fuchao {Wu}.
\newblock {L2-Net}: Deep learning of discriminative patch descriptor in
  euclidean space.
\newblock In {\em CVPR}, 2017.

\bibitem{umeyama1991least}
Shinji Umeyama.
\newblock Least-squares estimation of transformation parameters between two
  point patterns.
\newblock {\em TPAMI}, 1991.

\bibitem{wang2022improving}
Ziming Wang, Xiaoliang Huo, Zhenghao Chen, Jing Zhang, Lu Sheng, and Dong Xu.
\newblock Improving rgb-d point cloud registration by learning multi-scale
  local linear transformation.
\newblock In {\em ECCV}, 2022.

\bibitem{wilson20141dsfm}
Kyle Wilson and Noah Snavely.
\newblock Robust global translations with 1dsfm.
\newblock In {\em ECCV}, 2014.

\bibitem{yang2021self}
Heng Yang, Wei Dong, Luca Carlone, and Vladlen Koltun.
\newblock Self-supervised geometric perception.
\newblock In {\em CVPR}, 2021.

\bibitem{yew2020RPMNet}
Zi~Jian Yew and Gim~Hee Lee.
\newblock {RPM-Net: Robust Point Matching using Learned Features}.
\newblock In {\em CVPR}, 2020.

\bibitem{yi2018learning}
Kwang~Moo Yi, Eduard Trulls, Yuki Ono, Vincent Lepetit, Mathieu Salzmann, and
  Pascal Fua.
\newblock Learning to find good correspondences.
\newblock In {\em CVPR}, 2018.

\bibitem{zeng20163dmatch}
Andy Zeng, Shuran Song, Matthias Nie{\ss}ner, Matthew Fisher, Jianxiong Xiao,
  and Thomas Funkhouser.
\newblock {3DMatch: Learning local geometric descriptors from RGB-D
  reconstructions}.
\newblock In {\em CVPR}, 2017.

\bibitem{zhang1995robust}
Zhengyou Zhang, Rachid Deriche, Olivier Faugeras, and Quang-Tuan Luong.
\newblock A robust technique for matching two uncalibrated images through the
  recovery of the unknown epipolar geometry.
\newblock {\em Artificial intelligence}, 1995.

\bibitem{open3d}
Qian-Yi Zhou, Jaesik Park, and Vladlen Koltun.
\newblock {Open3D}: {A} modern library for {3D} data processing.
\newblock {\em arXiv}, 2018.

\bibitem{zhou2017unsupervised}
Tinghui Zhou, Matthew Brown, Noah Snavely, and David~G Lowe.
\newblock Unsupervised learning of depth and ego-motion from video.
\newblock In {\em CVPR}, 2017.

\end{thebibliography}
}

\end{document}